\documentclass[letterpaper]{article}
\usepackage{proceed2e}
\usepackage[margin=1in]{geometry}

\usepackage{times}

\usepackage[T1]{fontenc}
\usepackage[utf8]{inputenc}
\usepackage{bm}
\usepackage{amsmath}

\usepackage{amsthm}
\usepackage{amssymb}
\usepackage{graphicx,xspace,color}
\usepackage[unicode=true,
 bookmarks=false,
 breaklinks=false,pdfborder={0 0 1},backref=false,colorlinks=false]
 {hyperref}
\hypersetup{
 pdfstartview=FitH}

\makeatletter

  \theoremstyle{definition}
  \newtheorem{defn}{\protect\definitionname}
  \theoremstyle{plain}
  \newtheorem{assumption}{\protect\assumptionname}
\theoremstyle{plain}
\newtheorem{thm}{\protect\theoremname}
  \theoremstyle{definition}
  \newtheorem{problem}{\protect\problemname}
  \theoremstyle{plain}
  \newtheorem{prop}{Proposition}

\usepackage{bm,mathtools}
\usepackage{algorithm, algpseudocode,balance}
\usepackage[capitalise]{cleveref}
\usepackage{cite}

\newcommand{\To}{\textbf{to} }

\@ifundefined{showcaptionsetup}{}{%
 \PassOptionsToPackage{caption=false}{subfig}}
\usepackage{subfig}
\makeatother

  \providecommand{\assumptionname}{Assumption}
  \providecommand{\definitionname}{Definition}
  \providecommand{\problemname}{Problem}
\providecommand{\theoremname}{Theorem}
\usepackage{epstopdf}
\global\long\def\p{\operatorname{Pr}}

\global\long\def\s{\mathbf{s}}
\global\long\def\d{\mathrm{d}}
\global\long\def\t{\tau}
\global\long\def\a{\mathbf{A}}
\global\long\def\c{\mathbf{C}}
\global\long\def\dd{\mathbf{d}}
\global\long\def\h{\mathbf{H}}
\global\long\def\b{\mathbf{B}}
\global\long\def\lt{\operatorname{LowerTri}}
\global\long\def\u{\mathbf{u}}
\global\long\def\s{\mathbf{S}}
\global\long\def\D{\mathbf{D}}
\global\long\def\argmax{\operatorname{arg\,max}}
\global\long\def\argmin{\operatorname{arg\,min}}
\global\long\def\tt{\mathbf{t}}

\global\long\def\aa{\bm{\alpha}}
\global\long\def\g{\bm{\gamma}}
\global\long\def\x{{\bf x}}

\newcommand{\Alg}{{\textsc{Vine}}\xspace}
\newcommand{\AlgSA}{{\textsc{SimAnn}}\xspace}

\title{Submodular Variational Inference for Network Reconstruction}

\author{Lin Chen$^{1,2}$, Forrest W Crawford$^{2,3}$ \and Amin Karbasi$^{1,2}$ \\
$^{1}$Department of Electrical Engineering, 
$^{2}$Yale Institute for Network Science, $^{3}$Department of Biostatistics,
\\
Yale University
\\
\texttt{\{lin.chen,forrest.crawford,amin.karbasi\}@yale.edu}
}
\begin{document}
	

%
%
%
	
	%

\maketitle
\begin{abstract}
In real-world and online social networks, individuals 
receive and transmit information in real time. 
Cascading information transmissions (e.g.\ phone calls, text messages, social media posts)
may be understood as a realization of a diffusion process operating on the network.
 The process
 only traverses and thereby reveals  a limited portion of the edges.
The network reconstruction/inference problem 
is to estimate the unrevealed connections.
 Most existing approaches
derive a likelihood
and attempt to find the network topology maximizing
 the likelihood, yielding a highly intractable problem. 
 In this paper, we focus on the network reconstruction problem
for a broad class of real-world diffusion processes, exemplified by a network diffusion scheme called 
respondent-driven sampling (RDS).
 We prove that under  realistic and general models of network diffusion, the posterior distribution of an observed RDS realization is a Bayesian log-submodular model. We then propose \Alg
  , a novel, accurate, and computationally efficient variational 
inference algorithm, for the network reconstruction problem under this model. Crucially, we do not assume any particular probabilistic model for the underlying network.
 \Alg recovers any connected graph with high accuracy as 
shown by  our experimental results on real-life networks.
\end{abstract}

\section{Introduction}
The network reconstruction problem, also known as the network inference
problem \cite{
	gomez2010inferring,
daneshmand14netrate,kramer2009network,farajtabar2015back,
liben2007link,linderman2014discovering,
manuel13dynamic,
shandilya2011inferring,anandkumar2011topology,hanneke2009network,kim2011network},
arises naturally in a variety of scenarios and has been the focus of great
research interest. In the most general setting, we assume
there is an underlying unknown graph structure that represents
the connections between network subjects, and that 
we can only observe single or multiple diffusion processes over the
graph. Usually propagation of the diffusion process can only occur over network edges; 
however, there exist many hidden ties untraversed or unrevealed by the
diffusion processes, and the goal is to infer such hidden ties. 
This network reconstruction problem arises in several empirical topic areas:

\textbf{Blogosphere. }Millions of authors in the worldwide blogosphere
write articles every day, each triggering a diffusion process of reposts over
the underlying blog network structure.
The diffusion process initiated
by an article can be represented by a directed tree. The observed
data consist of multiple directed trees and it is of great interest
to understand the underlying structure of information flow
\cite{rodriguez2014uncovering}. Following inference of the network, 
researchers may apply community detection algorithms to,
 e.g.,
aggregate and further analyze blog sites of different political views.

\textbf{Online social networks. }Weibo is a Twitter-like microblogging
service in China \cite{gao2012comparative} where users 
 post
microblogs and
 repost those from other users they follow. 
 An explicit repost chain, which indicates the
sequence of users that a post passes through, is attached to each
repost on Weibo.
Similarly, each
post initiates a diffusion process. By observing several realizations
of diffusion processes, researchers seek to understand the underlying
social and information network structure.

\textbf{Respondent-driven sampling. }
Respondent-driven sampling (RDS)
is a chain-referral peer recruitment procedure that is widely used in epidemiology
for studying hidden and hard-to-reach human populations when random sampling 
is impossible \cite{heckathorn1997respondent}. 
RDS is commonly used in studies of 
men who have sex with men, homeless people, sex workers,
drug users, and other groups at high risk for HIV infection \cite{woodhouse1994mapping}.
An RDS recruitment process is also a diffusion process over
an unknown social network structure, and the diffusion tree (who recruited whom) 
is revealed by the observed
  process.  In addition, when a subject enters  the survey, she reports her total number of  acquaintances in the population, or graph-theoretically speaking, her degree in the underlying network.
%
Understanding the
underlying network structure is a topic of great interest to epidemiologists and sociologists 
who wish to study the transmission of infectious diseases, and the propagation of health-related
behaviors in the networks of high-risk groups \cite{crawford2016graphical}. However, in contrast to the aforementioned 
 scenarios where multiple diffusion realizations are available over the same
 network, in RDS we can only observe a single realization
  due to limited financial, temporal and human resources to conduct the experiments. As a result, network reconstruction from RDS data is particularly challenging and only heuristic algorithms are known. 
 Crawford \cite{crawford2016graphical} assumes that the recruitment time along any recruitment link is exponentially distributed and thus models RDS as a continuous-time diffusion process. Chen et al.~\cite{chen2015aaai} relaxes the requirement of exponentially distributed recruitment times and extends it to any distribution. Both works use a simulated-annealing-based heuristic in order to find the most likely configuration.

As a general strategy, for a particular diffusion
model, a likelihood function can be derived that measures the probability
of a diffusion realization.
 In this way, the network inference problem can be cast
as an optimization problem, in which the researcher seeks the topology 
that maximizes the likelihood.
 Unfortunately,
the derived likelihood functions
 are 
usually intractable for
efficient maximization with respect to the graph, and can be computationally prohibitive to evaluate. To address this challenge, approximate
solutions have been proposed as an efficient alternative \cite{gomez2010inferring,
	manuel13dynamic}. For instance, Gomez-Rodriguez et al.~\cite{gomez2010inferring}, instead of maximizing the likelihood,
 derived an alternative heuristic formulation by considering only the
most likely tree (still an NP-hard problem) rather than all possible propagation trees and showed how a greedy solution can find a near-optimal solution. 
It enjoys good empirical results when many realizations of the diffusion process can be observed.  
%

In this paper, we consider  the challenging instance of network inference where only one realization of the diffusion process is observed. As a motivating empirical example, we study the network reconstruction problem for RDS data 
  and propose \Alg (Variational Inference for Network rEconstruction), a computationally efficient variational inference algorithm. 
Our major contributions are summarized as follows.

%
%
%


\textbf{Proof of log-submodularity and a variational inference algorithm.} 
  We show that under a realistic model of RDS diffusion, the likelihood function is
log-submodular. Using variational inference methods, we approximate
the submodular function 
with  affine
 functions and obtain tight upper and lower bounds for the partition function.
We then estimate the most probable network configuration, which is
the maximizer of the likelihood,
as well as the marginal probability of each edge.

\textbf{Relaxation of constraints.} 
  The optimization problem of the RDS likelihood (as shown later) is constrained. First, the observed diffusion 
   results in a directed subgraph
    and the inferred network must be a supergraph of the diffusion process. Second, for each subject, 
  their degree in the reconstructed subgraph cannot exceed their total network degree.
   The first constraint is easy to incorporate while the second precludes efficient computation of partition functions of the likelihood (or any linear approximations). 
    We address this challenge by introducing penalty terms in the objective function. This way, the constrained reconstruction problem becomes unconstrained and admits the use of  variational methods.
%

\textbf{Flexibility for possibly inexact reported degrees.} 
  One may not assume that the reported degrees by recruited subjects are exact
    because subjects may not be able to accurately recall the number of people they know who are members of the target population. 
   We would like to note that the aforementioned relaxation of the second constraint
   allows for more flexibility of the possible mismatch of the reported degrees from the true ones by introducing an additional term that penalizes the deviation between the reported and true degrees, seeking to preserve the relative accuracy of the reported degrees.
  
\textbf{High reconstruction performance and time efficiency using a single realization of  RDS diffusion.} As shown by our experiments, \Alg achieves significantly higher inference performance while running orders of magnitude faster.  We should note that the very accurate inference is achieved based on the observation of a single diffusion realization. This is  in sharp contrast to previous work that assumes multiple diffusion realizations.

The rest of the paper is organized as follows. 
 In \cref{sec:Network-Reconstruction-for},
we focus on network reconstruction for RDS data and formulate
it
 as an optimization problem. We present our method in \cref{sec:proposed_method}. Experimental results
are presented in \cref{sec:Experiment}. All proofs are presented in \cref{sub:likeli,sub:logl,sub:proofthm,sub:lower,sub:sup}.
 Additionally,  
  we discuss the connection between RDS and other diffusion processes in \cref{sec:discuss}.


\section{Network Reconstruction for RDS Data}
\label{sec:Network-Reconstruction-for}

 \begin{figure*}[htb]
	\subfloat[\label{fig:exGR1}]{\includegraphics[width=0.175\textwidth]{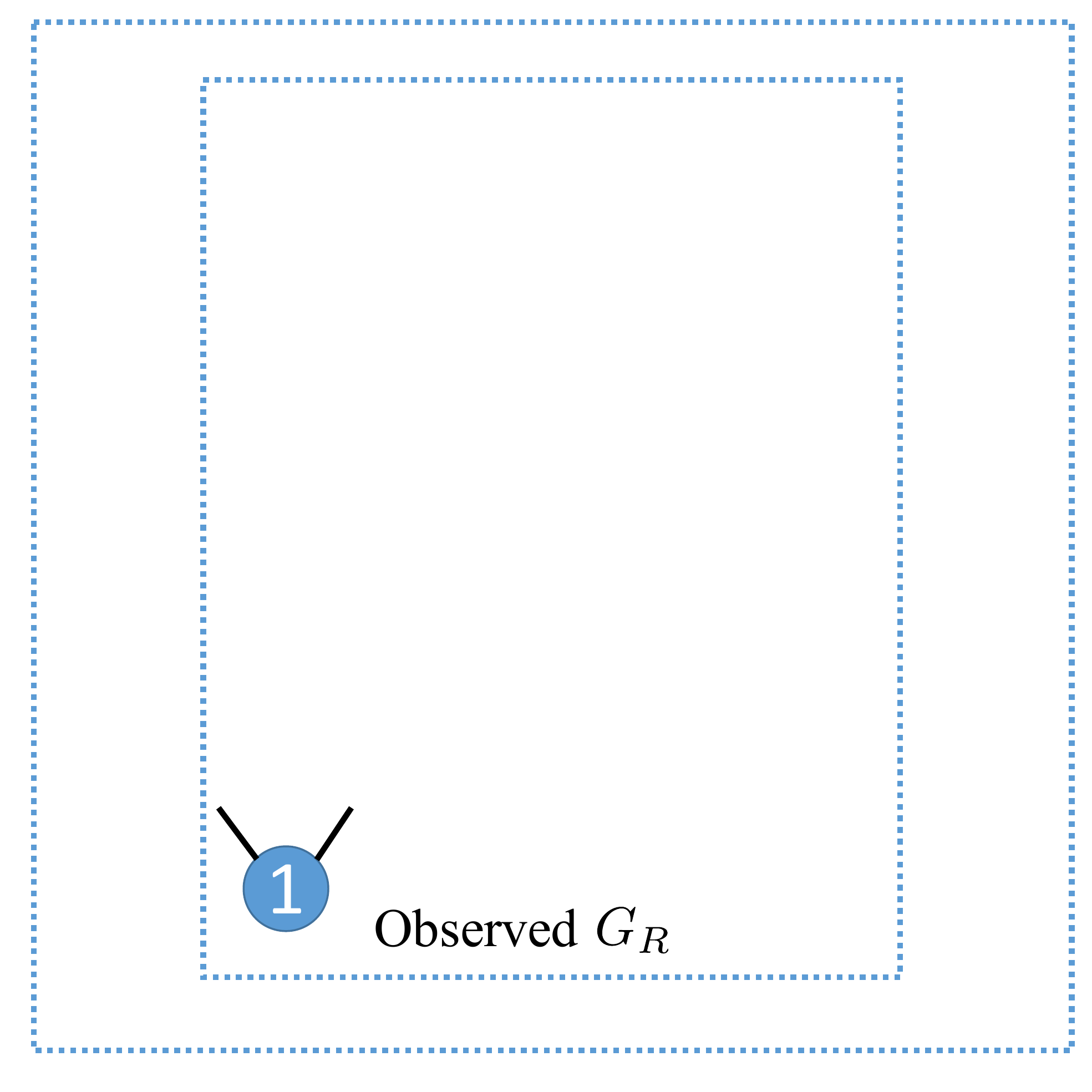}}
	\subfloat[\label{fig:exGR2}]{\includegraphics[width=0.175\textwidth]{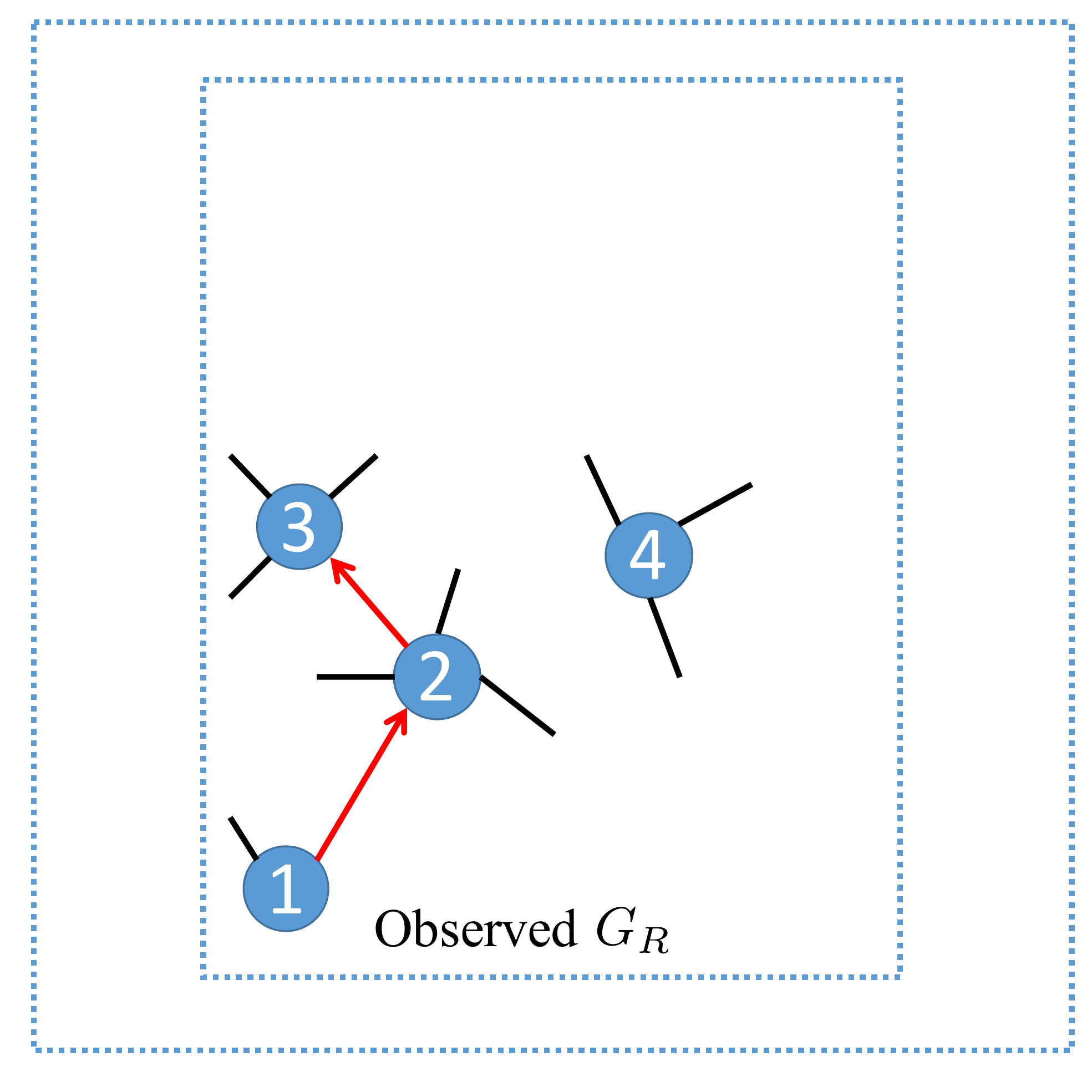}}	
	\subfloat[\label{fig:exGR3}]{\includegraphics[width=0.175\textwidth]{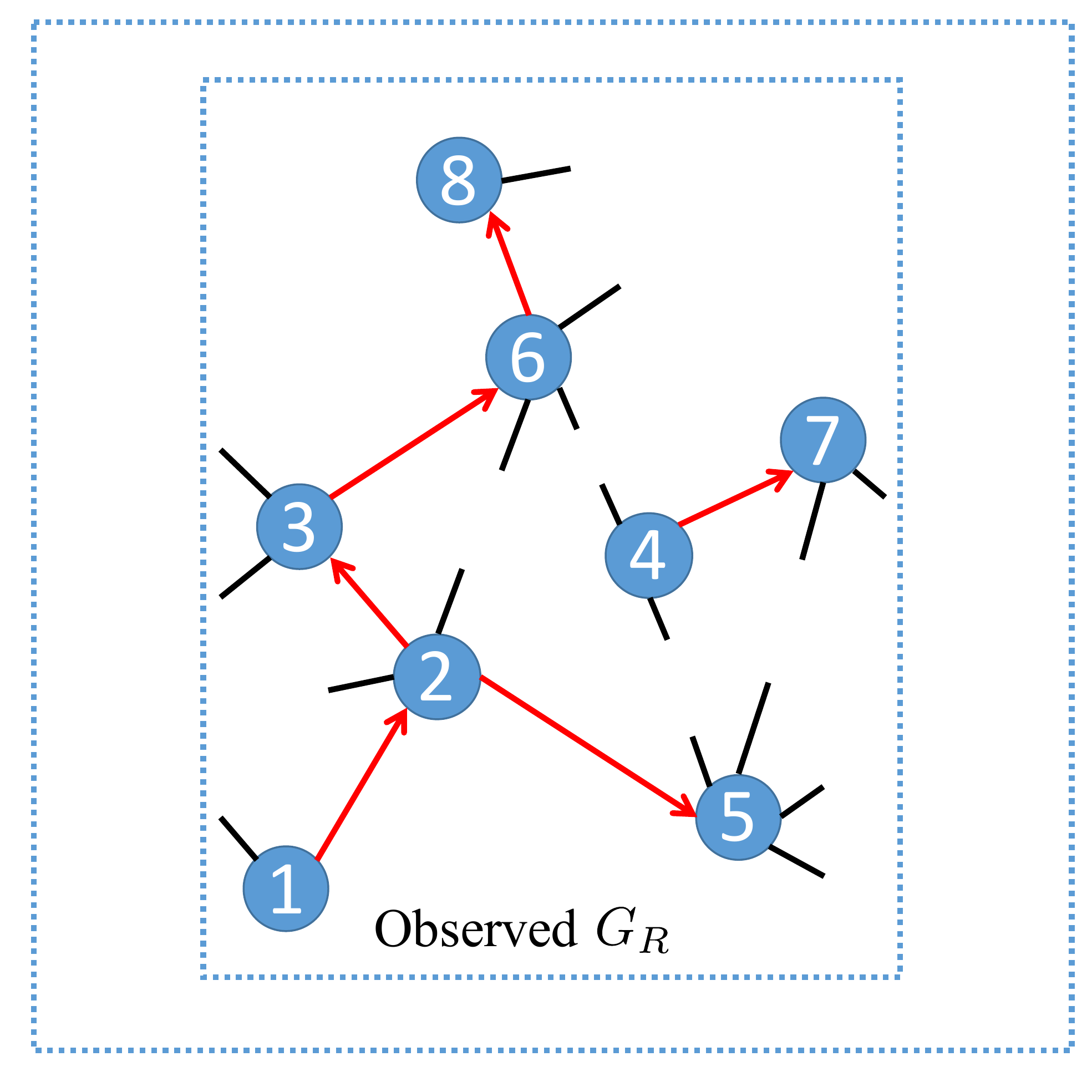}}
	\subfloat[\label{fig:exG}]{\includegraphics[width=0.175\textwidth]{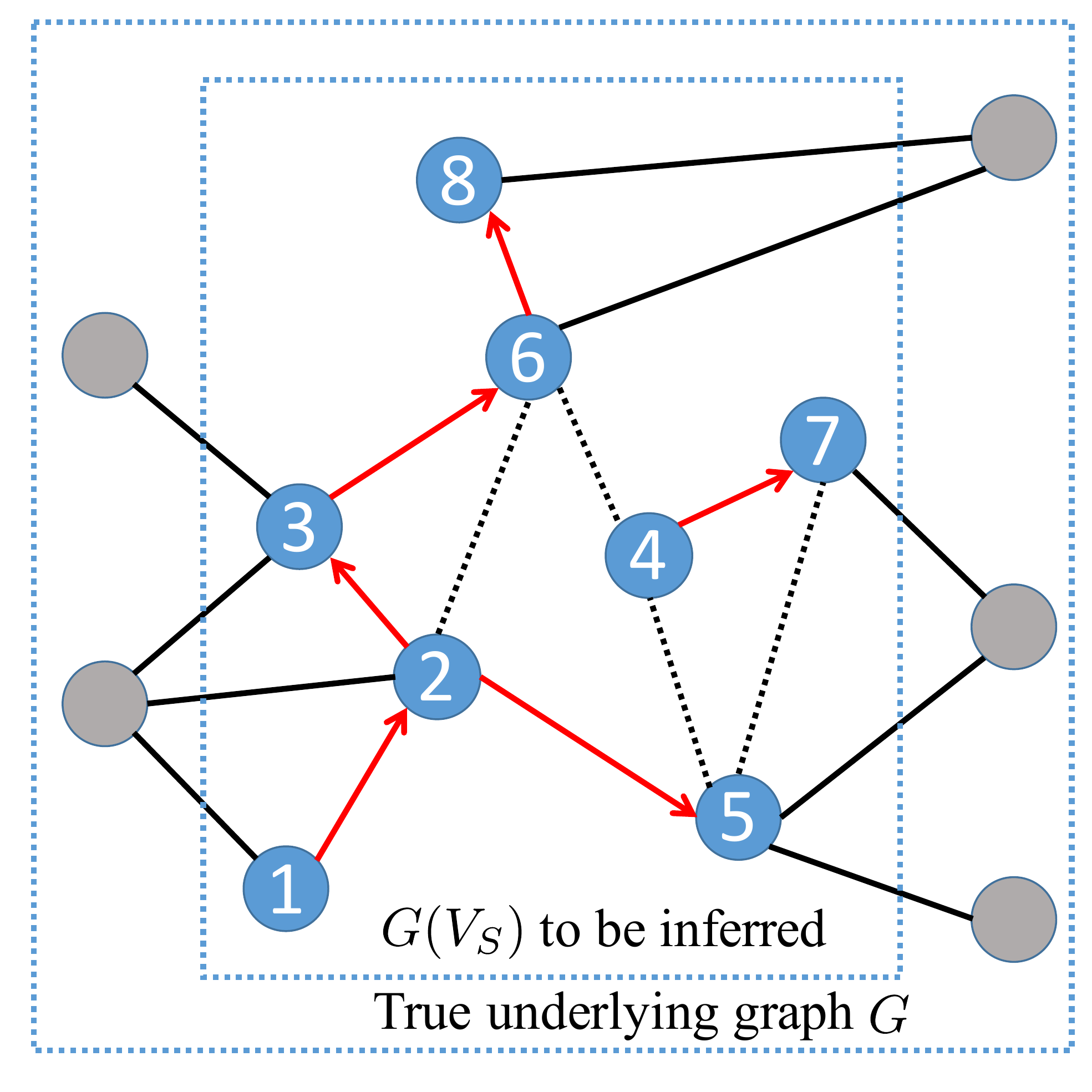} }	
	\centering
	\subfloat[\label{fig:d}]{\includegraphics[height=1in]{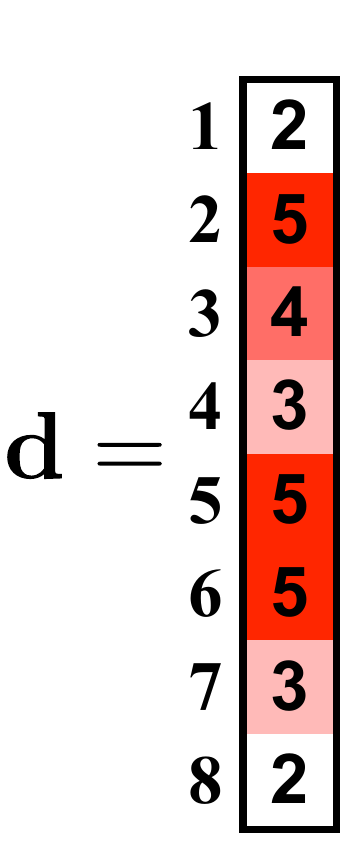}}
	\subfloat[\label{fig:t}]{\includegraphics[height=1in]{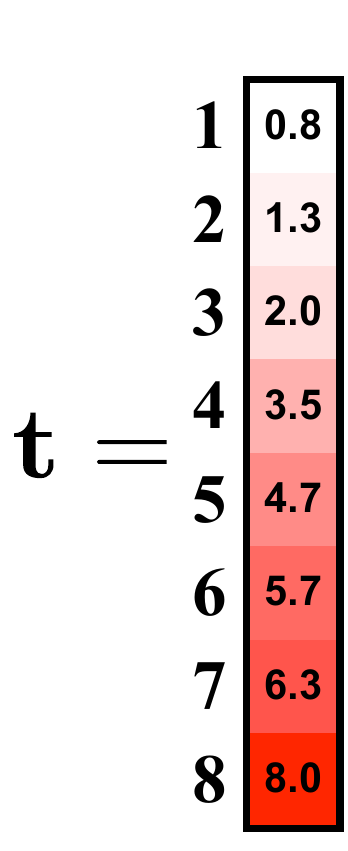}}
	\subfloat[\label{fig:c}]{\includegraphics[height=1in]{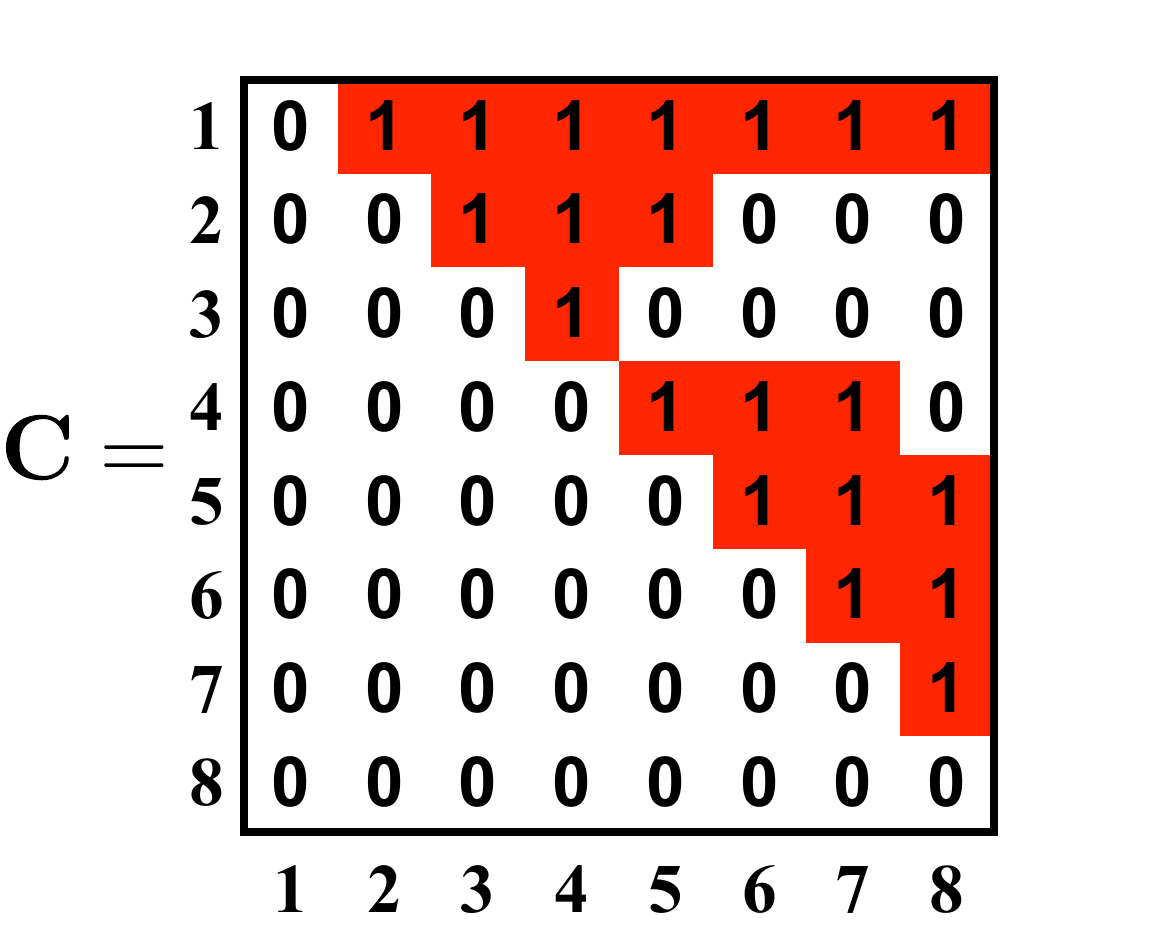}}
	\caption{
		\small 
		Example RDS recruitment with unobserved and observed data. 
		Nodes $ 1 $ and $ 4 $ are seed nodes chosen directly by the researchers; however, only node $ 1 $ is recruited at the beginning of the study.
		Every subject except nodes $3$ and $ 4 $ is given two coupons, while nodes $3$ and $ 4 $ are given only one. 
		\cref{fig:exGR1,fig:exGR2,fig:exGR3} show three snapshots of the RDS process at three different times.  \cref{fig:exGR1} shows the snapshot when the seed node $ 1 $ just enters the study due to direct recruitment by the researchers. Node $ 1 $ reports that her total degree is  $ 2 $; thus there are two pendant edges attached to her.
		\cref{fig:exGR2} presents the snapshot when the other seed node $ 4 $ is directly recruited by the researchers. Before node $ 4 $ is recruited, node $ 1 $ recruits node $ 2 $ and then node $ 2 $ recruits node $ 3 $. Red arrows denote the recruitment relation. Nodes $ 2 $, $ 3 $, and $ 4 $ report that their total degrees are $ 5 $, $ 4 $, and $ 3 $, respectively; thus there are $ 5 $, $ 4 $, and $ 3 $ pendant edges attached to them, respectively. Fig.~\ref{fig:exGR3} illustrates the snapshot at the end of the experiment.
		Fig.~\ref{fig:exG}  reveals the whole picture of the RDS process, containing the observable part (within the inner rectangle) and the unobservable part (outside the inner rectangle), where the inner rectangle denotes the sample and the outer rectangle denotes the entire population.
		The dashed lines denote the hidden ties to be inferred within the sample. The nodes outside the inner rectangle are marked in gray and they are the unsampled and thereby unobservable nodes.
		In Fig.~\ref{fig:d}, $\dd=(\dd_1,\dd_2,\dd_3,\ldots,\dd_n)'$ is the degree vector, where $\dd_i$ is the total degree of node $i$ in $G$. In Fig.~\ref{fig:t}, $\tt=(\tt_1,\tt_2,\tt_3,\ldots,\tt_n)'$ is the recruitment time vector, where $\tt_i$ is the recruitment time of node $i$. In Fig.~\ref{fig:c},  $\c$ is the coupon matrix; its $(i,j)$-entry is $ 1 $ if node $i$ has at least one coupon just before the $j$th recruitment event and is $ 0 $ otherwise. The observed data consist of $\mathbf{Y}=(\c,\dd,\tt,G_{R})$.\label{fig:exRDS} }
	
\end{figure*}

We use the following notational convention throughout this paper. The symbol $\mathbf{1}$ denotes the all-ones column vector.
If $f$ is a real-valued function and that $\mathbf{v}$
is a vector, then $f(\mathbf{v})$ is a vector of the same size as
vector $\mathbf{v}$, we denote the $i$-th entry of $f(\mathbf{v})$ by $f(\mathbf{v})_{i}$, and  $f(\mathbf{v})_{i}=f(\mathbf{v}_{i})$.
The transposes
of matrix $\a$ and vector $\mathbf{v}$ are written as $\a'$ and
$\mathbf{v}'$, respectively.

The objective of the RDS sampling method is to obtain a sample from a population for which random sampling is impossible. The network structure
of the underlying inaccessible population is modeled as an undirected
simple graph 
$G=(V,E)$, where
each vertex represents an individual
 and edges represent
the intra-population connections. The sample obtained via RDS is denoted by $ V_S \subseteq V$.
 Let $n=|V_S|$ be the number of subjects recruited into the study by the end of the RDS process.

In contrast to random sampling, RDS is a chain-referral process that operates by propagating a diffusion process on the edges of the target  
social network. Subjects enter the study one at a time.  The recruitment (diffusion) process is done  either by researchers directly or by other subjects already in the study prior to this new recruitment event. If a subject is recruited into the study by researchers directly,
she is called a \emph{seed}. Let $ M\subseteq V_S $ be the set of all seed nodes. 
  Note that seed nodes may not necessarily be recruited simultaneously; however, we need at least one seed that enters the study at the initial stage of the experiment in order to initiate the chain-referrals over the underlying network. We label the subjects $i=1,\ldots,n$ in the time-order they enter the study;  node $ i\in V_S $ is
the $i$-th subject that enters the study. When a subject enters the study (either via researchers directly or other subjects already in the study), she will be given several coupons
 to recruit other members
 (each recruitment costs one coupon).  Each coupon is marked with a unique ID that can be traced back to the recruiter.  
Subjects are given a reward for being interviewed and  recruiting other eligible subjects. 
The date/time of 
every subject's recruitment is recorded
and every subject reports their total number of 
acquaintances (their network \emph{degree}).
Let $\tt_{i}$ be the recorded recruitment
time of subject $i$ and $\dd_{i}$ be the reported degree of subject
$i$ in the population (in the graph $G$). 
Let the recruitment
time and degree vector  be $\tt=(\tt_{1},\tt_{2},\ldots,\tt_{n})'$ and
 $\dd=(\dd_{1},\dd_{2},\ldots,\dd_{n})'$, respectively.

Once a subject recruits another subject the (directed) link between them will be revealed. The direction simply indicates who recruited whom. Furthermore, any subject who has entered the study with a coupon may not re-enter the study with another coupon, and no participant may enter the study more than once. Thus no subject  can  recruit others already recruited and thereby already in the study. We can form a directed graph, called the recruitment graph $G_R=(V_R,E_R)$, that has the same vertex set as $ V_S $ and reflects the recruitment links; $ (i,j)\in E_R $ if and only if subject $ i $ directly recruits $ j $. The above requirements will result in a directed recruitment graph $ G_R $ being a disjoint union of rooted directed arborescences (a directed graph is a rooted directed arborescence with root $ r$ if for every vertex $v$, 
there exists a unique directed path from $ r $ to $ v $ )~\cite{gordon1989greedoid}, where the root
 corresponds to a seed node. Equivalently, an arborescence is a directed, rooted tree in which all edges point away from the root.  We illustrate an example of $ G_R $ in Fig.~\ref{fig:exG}, where the red links form the recruitment graph $ G_R $ and there are two disjoint arborescences with roots $ 1 $ and $ 4 $, respectively; the two roots correspond to the two seed nodes.

The first subject enters
the study at time $\tt_1$. At some time $t\geq \tt_1$, any
subject in the study who has at least one coupon (recall that each recruitment costs one coupon and that one cannot recruit any  subject without a coupon) and has at least
one acquaintance not in the study (i.e., has at least one  neighbor in $ G $ who is not already recruited) is called a \emph{recruiter} at
time $t$; accordingly, any subject who is not in the study and is connected to
at least one recruiter is called a \emph{potential recruitee} or a
\emph{susceptible subject }at time $t$; and the edge between a recruiter
and a potential recruitee is said to be \emph{susceptible }at time
$t$.  
Let $R(i)$ and $I(i)$ be the
recruiter set and potential recruitee set just before time $\tt_{i}$,
respectively. Similarly, If subject $ u $ is a recruiter just before time $ \tt_i $, then $ I_u(i) $ denotes the set of potential recruitees connected to subject $ u $ just before time $ \tt_i $, and if $ u $ is a potential recruitee just before time $ \tt_i $, then $ R_u(i) $ denotes the set of recruiters connected to subject $ u $ just before time $ \tt_i $.

In what follows, we model RDS as a continuous-time sto\-cha\-stic process on the edges of a hidden graph.  Our goal is to estimate the induced subgraph  connecting the sampled vertices $ G_S $.  To do this, we construct a flexible model for this process and derive its likelihood, conditional on an underlying graph.  The inference problem is to find the graph that maximizes this likelihood, subject to the constraint that the graph must be compatible with the observed degrees in the data. 
We start with making the following assumptions:
\begin{assumption}
Upon entering the study, each subject is given coupons and begins
to recruit other members (if any) immediately.
\end{assumption}
\begin{assumption}\label{asm:iid}
Inter-recruitment times 
between any recruiter and its potential
recruitees are i.i.d. continuous random variables with cumulative distribution
function (cdf) $D(t;\theta)$ 
parametrized by $\theta\in\Theta$. 
\end{assumption}
In fact, Assumption~\ref{asm:iid} can be relaxed to the case where inter-recruitment times
  are independent but not necessarily identically distributed. For simplicity we assume that they are i.i.d.

  If $W_{\theta}$ is a random variable with cdf $D(t;\theta)$, we
have $D(t;\theta)=\p\left[W_\theta \leq t\right]$ and let $D_{s}(t;\theta)=\p\left[W_{\theta}\leq t|W_{\theta}>s\right].$
We write $\rho_{s}(t;\theta)=\frac{\d D_{s}(t;\theta)}{\d t}$ for the conditional
probability density function (pdf). Let $S_{s}(t;\theta)=1-D_{s}(t;\theta)$
be the conditional survival function and $H_{s}(t;\theta)=\frac{\rho_{s}(t;\theta)}{S_{s}(t;\theta)}$
be the conditional hazard function. Recall that the set of all subjects
in the study is denoted by $V_{S}=\{1,2,3,\ldots,n\} \subseteq V$. The recruitment
graph $G_{R}=(V_{R},E_{R})$ has the same vertex set as $V_{S}$ 
and indicates who recruited whom: $(i,j)\in E_{R}$
if subject $i$ recruits subject $j$.
Note that subject $ i $ can recruit subject $ j $ only if there is an edge in the underlying network $ G $ that connects $ i $ and $ j $.
 The $n \times n$ coupon matrix $\c$ has a $1$ in entry 
$\c_{ij}$  if subject $i$ has
at least one coupon just before the $j$-th recruitment event $\tt_{j}$,
 and zero otherwise. In addition,
we define another $n\times n$ matrix $\a_{R}$, which is the adjacency
matrix of the undirected version of $G_{R}$, obtained
by replacing all directed edges with undirected edges.

\begin{assumption}
The observed data from an RDS process consists of $\mathbf{Y}=(\c,\dd,\tt,G_{R})$.
\end{assumption}

Our goal is to infer the induced subgraph $G(V_{S})$, denoted by $G_{S}=(V_{S},E_{S})$, which encodes all connections among
the subjects in the study. We also use $\a$ to denote the adjacency
matrix of
  $G_{S}$ and throughout this paper $\a$ and $G_{S}$
are used interchangeably. Obviously the undirectified version of $ G_R $ must be a subgraph of $ G_S $. Thus $ \a $ must be greater than or equal to $ \a_R $ entrywise; formally, \[ \a\geq\a_{R}\text{ (entrywise)}. \] This will be a constraint in the optimization problem specified later in Problem~\ref{prob:main}.
Fig.~\ref{fig:exRDS} shows an example 
of an
 RDS process including its 
 unobserved and observed parts.



Recall that $M$ denotes the set of seeds and let $\t(u;i)=\tt_{i-1}-\tt_{u}$.
The likelihood of the recruitment time series is given by
\begin{align}
& L(\mathbf{t}|\a,\theta)\nonumber\\
= & \prod_{i=1}^{n}\left(\sum_{u\in R(i)}|I_{u}(i)|H_{\t(u;i)}(\tt_{i}-\tt_{u};\theta)\right)^{1\{i\notin M\}}\nonumber\\
& \times \prod_{j\in R(i)}S_{\t(j;i)}^{|I_{j}|}(\tt_{i}-\tt_{j};\theta) \label{eq:likeliexpr}
\end{align}
\textbf{(The proof 
	of \cref{eq:likeliexpr} 
	is presented in \cref{sub:likeli})}. The above model was originally derived in~\cite{chen2015aaai}.

We can represent the log-likelihood in a more compact way using linear algebra. Prior to this, we need some notation.
Let $\mathbf{m}$ and $\u$ be column vectors of size $n$ such that
$\mathbf{m}_{i}=1\{i\notin M\}$ and $\u_{i}$ is the number of pendant
edges of subject $i$ (
the reported total degree of subject $i$
minus the number of its neighbors in $G_{S}$), i.e., \[ \u_i =\dd_i-\left|\{j\in V_S:(i,j)\in E_S\}\right|, \] and let $\h$ and
$\s$ be $n\times n$ matrices, defined as $\h_{ui}=H_{\t(u;i)}(\tt_{i}-\tt_{u};\theta)$
and $\s_{ji}=\log S_{\t(j;i)}(\tt_{i}-\tt_{j};\theta).$ Furthermore,
we form matrices $\b=(\c\circ\h)$ and $\D=(\c\circ\s)$, where $\circ$
denotes the Hadamard (entrywise) product. We let 
\begin{eqnarray*}
\beta&=&\log(\b'\u+\lt(\a\b)'\cdot\mathbf{1}),\\
\delta&=&\D'\u+\lt(\a\D)'\cdot\mathbf{1},
\end{eqnarray*}
%
 where the log of a vector is taken entrywise and $\lt(\cdot)$ denotes the lower triangular part (diagonal elements
inclusive) of a square matrix. 
Then
 the log-likelihood
can be written as
\begin{equation}
l(\mathbf{t}|\a,\theta)=\mathbf{m}'\beta+\mathbf{1}'\delta \label{eq:loglikelihood}
\end{equation}
\textbf{(The proof of \cref{eq:loglikelihood} is presented in \cref{sub:logl})}. To adopt a Bayesian approach, we consider maximizing the joint posterior distribution 
\[
\p\left(\a,\theta|\tt\right)\propto L(\tt|\a,\theta)\pi(\a)\phi(\theta),
\]
where $\pi$ and $\phi$ are the prior distribution of $\a$ and $\theta$. The network inference problem of the RDS data is reduced to maximization of $\p\left(\a,\theta|\tt\right)$. 
Our main observation in this paper is that the log-likelihood function is submodular, which opens the possibility of rigorous analysis and variational inference.

If we assume that the reported degrees
of subjects are exact, then the vector $\u$ should be set to $\dd-\a\cdot\mathbf{1}$
and it must be non-negative entrywise.
However, in practice, the reported degree of a subject may be an
approximation, but we assume
the true degree does not deviate excessively from the reported degree. 
To be 
more flexible, 
we allow $\u$ to be any non-negative integer-valued
vector. In this case, the true degree vector will be $\dd_{\mathrm{true}}=\u+\a\cdot\mathbf{1}$.
We penalize it if $\dd_{\mathrm{true}}$ deviates from $\dd$ excessively.
To be precise, we define the prior distribution $\pi(\a)$ as
\begin{equation}
\pi(\a)\propto\exp(-\psi(\max\{\u+\a\cdot\mathbf{1}-\dd,\mathbf{0}\})),\label{eq:pi}
\end{equation}
where $\max$ is conducted entrywise
and $\psi$ is a multivariate ($n$-dimensional) convex function
and non-decreasing in each argument whenever this argument is non-negative. We can now formulate our inference problem as an optimization problem.
\begin{problem}\label{prob:main}
Given the observed data $\mathbf{Y}=(\c,\dd,\tt,G_{R})$, we seek
an $n\times n$ adjacency matrix (symmetric, binary and zero-diagonal)
and a parameter value $\theta\in\Theta$ that
\[
\begin{array}{cc}
\mbox{maximizes} & L(\tt|\a,\theta)\pi(\a)\phi(\theta)\\
\mbox{subject to} & \a\geq\a_{R}\text{ (entrywise)}.
\end{array}
\]
Problem~\ref{prob:main} can be solved 
by  maximizing the likelihood with respect to $ \theta $ and $ \a $ alternately.
 We set an initial guess $ \theta_1 $ for the parameter $ \theta $. In the $ \tau $-th iteration ($ \tau \geq 1 $), setting $ \theta = \theta_\tau $ in Problem~\ref{prob:main}, we optimize the objective function over $ \a $ (this step is called the $ \a $-step), denoting the maximizer by $ \a_\tau $; then setting $ \a = \a_\tau $ in Problem~\ref{prob:main}, we optimize the objective function over $ \theta $ (this step is called the $ \theta $-step), denoting the maximizer by $ \theta_{\tau+1} $. 
 The interested reader is referred to Algorithm~1 in~\cite{chen2015aaai}.
 Note that the parameter space $ \Theta $ is usually a subset of the Euclidean space. The optimization problem given $ \a $ in the $ \theta $-step can be solved with off-the-shelf solvers. As a result, we focus on the $ \a $-step; equivalently, we study how to solve Problem~\ref{prob:main} assuming that $ \theta $ is known.

\end{problem}

\section{Proposed Method}
\label{sec:proposed_method}

We now present a network reconstruction algorithm, based on submodular variational inference, for respondent-driven sampling data. This method is referred to as \Alg in this paper. We first introduce the definition of a submodular function~\cite{jegelka2011fast,iyer2015polyhedral}. 
\begin{defn}
A pseudo-Boolean function $f:\{0,1\}^{p}\to\mathbb{R}$ is
 \emph{submodular }if 
  $\forall {\bf x},{\bf y}\in\{0,1\}^{p}$,
we have
 $f({\bf x})+f({\bf y})\geq f({\bf x}\wedge{\bf y})+f({\bf x}\vee{\bf y})$,
where $\wedge$ and $\vee$ denote entrywise logical conjunction and
disjunction, respectively.
\end{defn}
We can trivially identify the domain $\{0,1\}^{p}$ with $2^{[p]}$,
the power set of $[p]=\{1,2,3,\ldots,p\}$. Thus a pseudo-Boolean
function $f$ can also be viewed as a set function $2^{[p]}\to\mathbb{R}$.
We will view $f$ from these two perspectives interchangeably throughout
this paper. If we view $f$ as a set function, it is submodular if
for every subset $X,Y\subseteq[p]$, we have $f(X)+f(Y)\geq f(X\cap Y)+f(X\cup Y)$.
An equivalent definition is that $f$ is submodular if for every $X\subseteq Y\subseteq[p]$
and $x\in[p]\setminus Y$, we have $f(X\cup\{x\})-f(X)\geq f(Y\cup\{x\})-f(Y)$;
this is also known as the ``diminishing returns'' property because the marginal
gain when an element is added to a subset is no less than the marginal
gain when it is added to its superset.

A pseudo-Boolean or set function $f$ is 
\emph{log-submodular
}if $\log(f)$ is submodular;
 it is 
  \emph{modular} if $f({\bf x})=\sum_{i=1}^{p}f_{i}{\bf x}_{i}$
(if viewed as a pseudo-Boolean function) or equivalently $f(X)=\sum_{i\in X}f_{i}$ (if viewed as a set function),
where $f_{i}\in\mathbb{R}$ is called the weight of the element $i$.
It is \emph{affine} if $f({\bf x})=s({\bf x})+c$, where
$s$ is modular
 and $c$ is some fixed real number;
similarly, it is
 \emph{log-affine
}if $\log(f)$ is affine.

\subsection{Removal of constraint in Problem~\ref{prob:main}}\label{sub:removal}
The formulation of Problem~\ref{prob:main} makes clear that the network reconstruction problem is
a constrained optimization problem.
Recall that we have two constraints. One is that the reconstructed subgraph should contain all edges already revealed by the RDS process. This constraint is natural since if a direct recruitment occurs between two subjects, then they must know each other in the underlying social network. The other constraint is that the degree of the subject in the reconstructed subgraph must be bounded by the degree that this subject reports.
In this section, we  remove these two constraints and cast it into an unconstrained problem.
The first constraint is easy to remove by considering only the edges unrevealed by the RDS process. The final objective function results from replacing the second constraint with a penalty function to allow for some room for the deviation of the degree in the inferred subgraph from the reported degree. After relaxing the two constraints, we turn Problem~\ref{prob:main} into an unconstrained problem and make submodular variational inference (to be discussed later in Section~\ref{sub:varinf}) possible.

 Specifically, the first constraint requires some entries of $\a$ to be $1$;  if
the $(i,j)$-entry of $\a_{R}$ (denoted by $\a_{R}^{ij}$) is $1$,
so is $\a$. Only the rest of the entries of $ \a $ can either be $ 0 $ or $ 1 $ and are the free entries. We collect the free entries in a binary
vector $$\aa=(\a_{ij}:1\leq i<j\leq n,\a_{R}^{ij}=0)$$ and view $L(\tt|\a,\theta)\pi(\a)\phi(\theta)$
as a function of $\aa$. In fact, there is a one-to-one correspondence between $ \a $ and $ \aa $. In this way, we remove the constraint $ \a \geq \a_R $.
Now we discuss how to relax the second constraint (the degree constraint) to make small deviation from the (usually approximate) reported degree possible.

\textbf{Representing $ \u $ as a binary vector and thereby the objective function as a pseudo-Boolean function.} We notice that $L(\tt|\a,\theta)\pi(\a)\phi(\theta)$
is also a function of $\u$; however, $ \u $ is an integer-valued vector rather than a binary vector.
We consider representing $ \u $ as a binary vector and thereby casting $L(\tt|\a,\theta)\pi(\a)\phi(\theta)$  into a pseudo-Boolean function. We observe that
$\u$ is bounded entrywise; to be precise, $\forall1\leq i\leq n$,
$0\leq\u_{i}\leq u_{\text{max}}$, where $u_{\text{max}}=\max_{1\leq i\leq n}(\dd-\a\cdot\bm{1})_{i}$.
We can form an $n\times\left\lceil \log_{2}u_{\text{max}}\right\rceil $
matrix $\bm{\mu}$ such that the $i$-th row of $\bm{\mu}$ is the
binary representation of $\u_{i}$; formally, 
\begin{equation}
\u=\bm{\mu}\cdot\left(\begin{array}{ccccc}
2^{0} & 2^{1} & 2^{2} & \ldots & 2^{\left\lceil \log_{2}u_{\text{max}}\right\rceil -1}\end{array}\right)'.\label{eq:u-mu}
\end{equation}

In this way, we represent $L(\tt|\a,\theta)\pi(\a)\phi(\theta)$ as
a pseudo-Boolean function of $\aa$ and $\bm{\mu}$. Let $\g=(\aa,\bm{\mu})$
and define $\tilde{L}(\g)=L(\tt|\a,\theta)\pi(\a)\phi(\theta).$ Therefore
$\tilde{L}$ is a pseudo-Boolean function of $\g$, whose dimension is 
$
N=\sum_{1\leq i<j\leq n}1_{\{\a_{R}^{ij}=0\}}+n\times\left\lceil \log_{2}u_{\text{max}}\right\rceil .
$
%
The  likelihood function $\tilde{L}(\g)$ defines a
 probability measure over
$\{0,1\}^{N}$, 
$\p(\g)=\tilde{L}(\g)/\tilde{Z} $, 
where  $\tilde{Z}=\sum_{\g\in \{0,1\}^N }\tilde{L}(\g)$ is the normalizing constant.

\subsection{Submodularity of log-likelihood function}\label{sub:subm}

 Theorem \ref{thm:log-submodular} below
shows that $\tilde{L}(\g)$ is log-submodular. We know that a submodular function can be approximated by affine functions from above and below. Due to the log-sub\-mo\-du\-larity of $ \tilde{L}(\g) $, it can be approximated by two log-affine functions from above and below. The partition function of a probability distribution proportional to a log-affine function can be computed in a closed form; therefore this distribution can be computed exactly and we have upper and lower bounds of $ \p(\g) $; we can conduct variational inference via the two bounds. We will elaborate on this in Section~\ref{sub:varinf}.
\begin{thm}[\textbf{Proof in Appendix~\ref{sub:proofthm}}]
\label{thm:log-submodular}  The function $\tilde{L}(\g)$ is log-submodular;
{equivalently}, there exists a submodular function $\tilde{F}(\g)$ such that $\tilde{L}(\g)=\exp \tilde{F}(\g)$
for every $\g\in\{0,1\}^{N}$.
\end{thm}

\textbf{Normalizing $ \tilde{L}(\g) $ into $ L(\g) $.} 
Ideally, we want a submodular function $F$ to be \emph{normalized}; i.e., $F(\bm{0})=0$, or equivalently $F(\varnothing) = 0$ if viewed as a set function. Thus we define $F(\g)\triangleq  \tilde{F}(\g) - \tilde{F}(\bm{0});$ this way $F$ is a normalized submodular function (it is a submodular function minus some constant). In addition, we define $L(\g) = \exp (F(\g))$ and we have $L(\g) = \exp (\tilde{F}(\g) - \tilde{F}(\bm{0})) = \exp(- \tilde{F}(\bm{0})) \tilde{L}(\g).$ Note that the probability measure is proportional to $\tilde{L}(\g)=L(\tt|\a,\theta)\pi(\a)\phi(\theta)$ (up a constant factor) and that $L(\g)$ only differs from $\tilde{L}(\g)$ up to a constant factor. Therefore the probability measure remains proportional to  $L(\g)$, thus  $L(\g)$ is also a likelihood function and defines the same
 probability measure over
$\{0,1\}^{N}$ as $\tilde{L}(\g)$ does.  As a result, the probability measure can be expressed as
$\p(\g)=\frac{1}{Z}\exp (F(\g)),$
where $Z=\sum_{\g\in \{0,1\}^N}\exp(F(\g))$ is the normalizing constant,
or the \emph{partition function}.


\subsection{Variational inference}\label{sub:varinf}
Using a variational method
lets us  consider
bounding $F(\g)$ from above and from below with affine functions.
 We want to find modular functions $s_{u}$
and $s_{l}$ and two real numbers $c_{u}$ and $c_{l}$ such that
$s_{l}(\g)+c_{l}\leq F(\g)\leq s_{u}(\g)+c_{u}$ for all $\g\in \{0,1\}^N$.
If this holds for all $\g\in \{0,1\}^N$, then we have the inequality between 
log-partition functions: $\sum_{\g\in \{0,1\}^N}\exp({s_{l}(\g)+c_{l}})\leq\sum_{\g\in \{0,1\}^N}\exp{F(\g)}=Z\leq\sum_{\g\in \{0,1\}^N}\exp({s_{u}(\g)+c_{u}}).$
We define the partition function of the affine function $s(\g)+c$
as $Z(s,c)\triangleq\sum_{\g\in \{0,1\}^N}\exp(s(\g)+c).$ Using this notation,
we have $Z(s_{l},c_{l})\leq Z\leq Z(s_{u},c_{u}).$ Note that from
this we may also obtain the bounds for the marginal probability for
each element $i\in[N]$. To be precise, if $\g$ is sampled from the
distribution $\p(\g)=\exp(F(\g))/Z$, then the marginal probability $\p(i\in\g)$
satisfies
$
\frac{s_{l}(\{i\})+c_{l}}{Z(s_{u},c_{u})}\leq\p(i\in\g)\leq\frac{s_{u}(\{i\})+c_{u}}{Z(s_{l},c_{l})}.
$
We may also use $s_{u}(\g)+c_{u}$ or $s_{l}(\g)+c_{l}$ as a surrogate
for $F$ and make inference via these two
 affine
functions.

Suppose that we already have two affine functions $s_{u}(\g)+c_{u}$ and $s_{l}(\g)+c_{l}$
bounding $F(\g)$ from above and below. By Lemma 1 in \cite{djolonga14variational},
the log-partition function for $s(\g)+c$ in the unconstrained case
is $\log Z'(s,c)\triangleq\log\sum_{\g\in\{0,1\}^{N}}\exp({s(\g)+c})=c+\sum_{i=1}^{N}\log(1+\exp{ s_{i}}),$
where $s_{i}=s(\{i\})$ is the weight of element $i$.
Thus we have
$
Z'(s_{l},c_{l})\leq Z=\sum_{\g\in\{0,1\}^{N}}L(\g)=\sum_{\g\in\{0,1\}^{N}}\exp{F(\g)}\leq Z'(s_{u},c_{u}).
$

So our goal is to find the upper- and lower-bound affine functions.

\textbf{Lower-bound affine function.}
 We  define
\[
v_{j}=\argmax_{k\in[N]\setminus V_{j}}\left(F(V_{j}\cup\{k\})-F(V_{j})\right),
\]
and
\[
s_{v_{j}}^{g}=\max_{k\in[N]\setminus V_j}\left(F(V_{j}\cup\{k\})-F(V_{j})\right),
\]
 where $ V_1=\varnothing $,  $V_{j}=\{v_{1},v_{2},\ldots,v_{j-1}\}$, and  $ 1\leq j \leq N $.
Then we have the affine function $s^{g}$ that assigns to $i\in[N]$ a
weight of $s_{i}^{g}$. 
Let $s_{l}(\g)=s^{g}(\g)$ and $c_{l}=0$.

\begin{prop}[\textbf{Proof in Appendix~\ref{sub:lower}}]\label{prop:lower} 
The affine function $s^g$ is a lower-bound function of the submodular function $F$; for all $ \g\in \{0,1\}^N$, $s^g(\g)\leq F(\g)$.
\end{prop}


\textbf{Upper-bound affine function.} We may find an upper-bound affine function for a submodular function
via its supergradients. The set of supergradients of a submodular
function $F$ at ${\bf x}\in\{0,1\}^{N}$ \cite{iyer2013-fast-submodular-semigradient}
is defined as
\begin{multline} \partial^{F}({\bf x})=\{s\text{ is modular}:\\
\forall{\bf y}\in\{0,1\}^{N},F({\bf y})\leq F({\bf x})+s({\bf y})-s({\bf x})\}.\nonumber
\end{multline}
If a modular function $s$ is a supergradient of $F$ at ${\bf x}$,
then the affine function $s(\g)+(F(\x)-s({\bf x}))$ is an upper bound
of $F(\g)$. The corresponding log-partition function is $Z'_{\x}(s)=Z'(s,F(\x)-s(\x)).$

\begin{algorithm}[htb]
	\caption{\Alg\label{alg:varinf}}
	\begin{algorithmic}[1]
		\Require observed data $\mathbf{Y}=(\c,\dd,\tt,G_{R})$ 
		\Ensure inferred adjacency matrix $ \hat{\a} $
		\Function{GetLowerBoundAffineFunction}{}
		\State $ V_1\gets \varnothing $
		\For{$ j\gets 1 $ \To $ N $}
		\State $ v_{j}\gets \argmax_{k\in[N]\setminus V_{j}}\left(F(V_{j}\cup\{k\})-F(V_{j})\right) $
		\State $ s_{v_{j}}^{g}\gets\max_{k\in[N]\setminus V_j}\left(F(V_{j}\cup\{k\})-F(V_{j})\right) $
		\State $ V_{j+1}\gets V_j \cup \{v_j\} $
		\EndFor
		\State \Return{affine function $ s_l(\g) = \sum_{j=1}^{N} s^g_j \g_j $}
		\EndFunction
		\Statex
		\Function{GetUpperBoundAffineFunction}{}
		\State $m_i\gets \log\left(1+e^{-\Delta_{i}F([N]\setminus\{i\})}\right)-\log\left(1+e^ {F(\{i\})}\right)$
		\State Define $ m(\x) = \sum_{i=1}^{N} m_i \x_i $
		\State $ \x \gets \argmin_{\x} (F(\x)+m(\x)) $
		\State $ S_{\x} = \{ \hat{s}_{{\bf x}},
		\check{s}_{{\bf x}} ,\bar{s}_{{\bf x}}  \} $
		\State $ s_u \gets \argmin_{s\in S_{\x}} Z'_{\x}(s) $
		\State \Return{affine function $ s_u(\g)+(F(\x)-s_u(\x)) $}
		\EndFunction
		\Statex
		\Function{VariationalInference}{}
		\State $ s\gets \Call{GetUpperBoundAffineFunction} $ or $ \Call{GetLowerBoundAffineFunction} $
		\State Select threshold $ \zeta\in[\min_{i}s_{i}^{\alpha},\max_{i}s_{i}^{\alpha}] $
		\State $ \bm{\alpha}_{i}\gets 1_{\{s_{i}^{\alpha}\geq\zeta\}}$
		\State Obtain the inferred adjacency matrix $ \hat{\a} $ from $ \bm{\alpha} $
		\EndFunction
	\end{algorithmic}
\end{algorithm}

We consider three
families of supergradient, which are grow ($\hat{s}_{{\bf x}}$),
shrink ($\check{s}_{{\bf x}}$) and bar ($\bar{s}_{{\bf x}}$) supergradients
at $\x$.
 Let us view $F(\g)$ as a set function where $\g\subseteq[N]$.
We define $\Delta_{i}F(\g)=F(\g\cup\{i\})-F(\g),$ where $i\in[N]$.
These three supergradients are defined as follows. If $j\in\x$, then
$\hat{s}_{\x}(\{j\})=\check{s}_{\x}(\{j\})=\Delta_{j}F(\x\setminus\{j\})$ and $ \bar{s}_{\x}(\{j\})=\Delta_{j}F([N]\setminus\{j\}) $
if $j\notin\x$, then $\hat{s}_{\x}(\{j\})=\Delta_{j}F(\x)$ and $\check{s}_{\x}(\{j\})=\bar{s}_{\x}(\{j\})=F(\{j\}).$

\begin{prop}[\textbf{Proof in Appendix~\ref{sub:sup}}]\label{prop:sup}
The modular functions $\hat{s}_{\x}$, $\check{s}_\x$ and $\bar{s}_\x$ are supergradients of the submodular function $F$ at $\x$.
\end{prop}

Define the modular function
 $$m(\{i\})=\log\left(1+e^{-\Delta_{i}F([N]\setminus\{i\})}\right)-\log\left(1+ e^{F(\{i\})}\right).$$
 By Lemma 4 in \cite{djolonga14variational},
we know that these two optimization problems are equivalent:
\[
\min_{\x}\log Z'_{\x}(\bar{s}^{\x})\equiv\min_{\x}F(\x)+m(\x).
\]
The right-hand side is an unconstrained submodular minimization problem,
which can be solved efficiently~\cite{jegelka2011fast}. By solving this problem, we obtain a supergradient $\bar{s}^{\x}$
at $\x$ and thus know its partition function $Z'_{\x}(\bar{s}^{\x})$. Then
we compute the partition function of grow and shrink supergradients
at $\x$ and let $s_{u}$ be the one with the smallest partition function.
Thus the upper-bound affine function is $ s_u(\g)+(F(\x)-s_u(\x)) $.

When we have an affine approximation for the submodular function,
we may make inference via the affine function. Suppose that the affine
function is $s(\g)+c$, which can be either an upper-bound or lower-bound
approximation. Recall that $\g=(\bm{\alpha},\bm{\mu})$ and 
$
s(\g)=\sum_{i=1}^{N_{1}}s_{i}^{\alpha}\bm{\alpha}_{i}+\sum_{i=1}^{n}\sum_{j=1}^{N_{2}}s_{ij}^{\mu}\bm{\mu}_{ij},
$
where $N_{1}=\sum_{1\leq i<j\leq n}1_{\{\a_{R}^{ij}=0\}},$ $N_{2}=\left\lceil \log_{2}u_{\text{max}}\right\rceil, $
and $s_{i}^{\alpha},s_{ij}^{\mu}\in\mathbb{R}$.

 We can select a threshold
$\zeta\in[\min_{i}s_{i}^{\alpha},\max_{i}s_{i}^{\alpha}]$ and obtain
$\bm{\alpha}(\zeta)$ by thresholding, $ \bm{\alpha}(\zeta)_{i}=1_{\{s_{i}^{\alpha}\geq\zeta\}}$.
Thus we obtain an inferred adjacency matrix $\hat{\a}(\zeta)$ from $\bm{\alpha}(\zeta)$.
The proposed method \Alg is presented in \cref{alg:varinf}.

\section{Experiment}
\label{sec:Experiment}

\begin{figure}[t]
	\subfloat[\label{fig:ex}]{
		\includegraphics[width=0.24\textwidth]{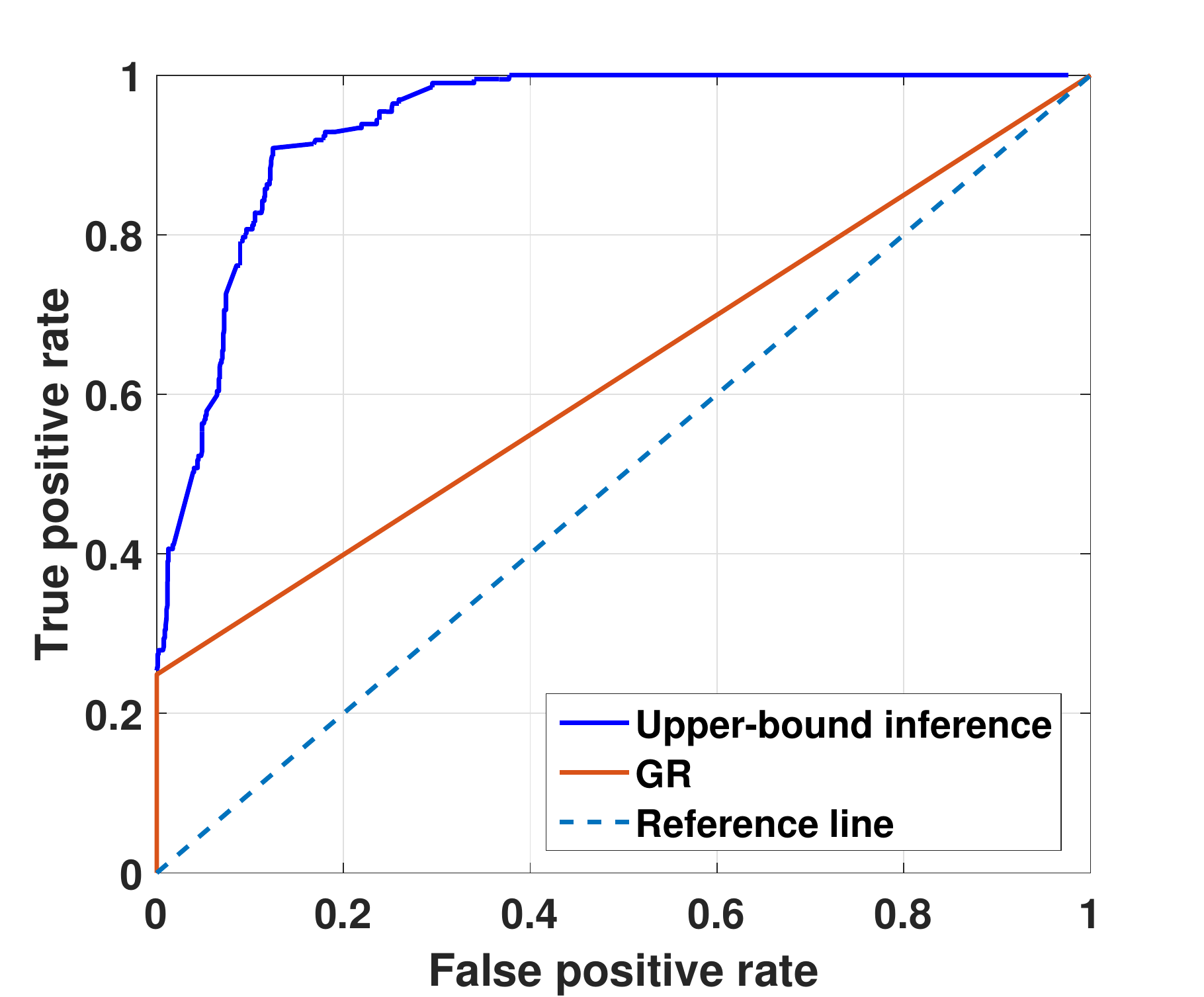}
	}\subfloat[\label{fig:ex_low}]{
		\includegraphics[width=0.24\textwidth]{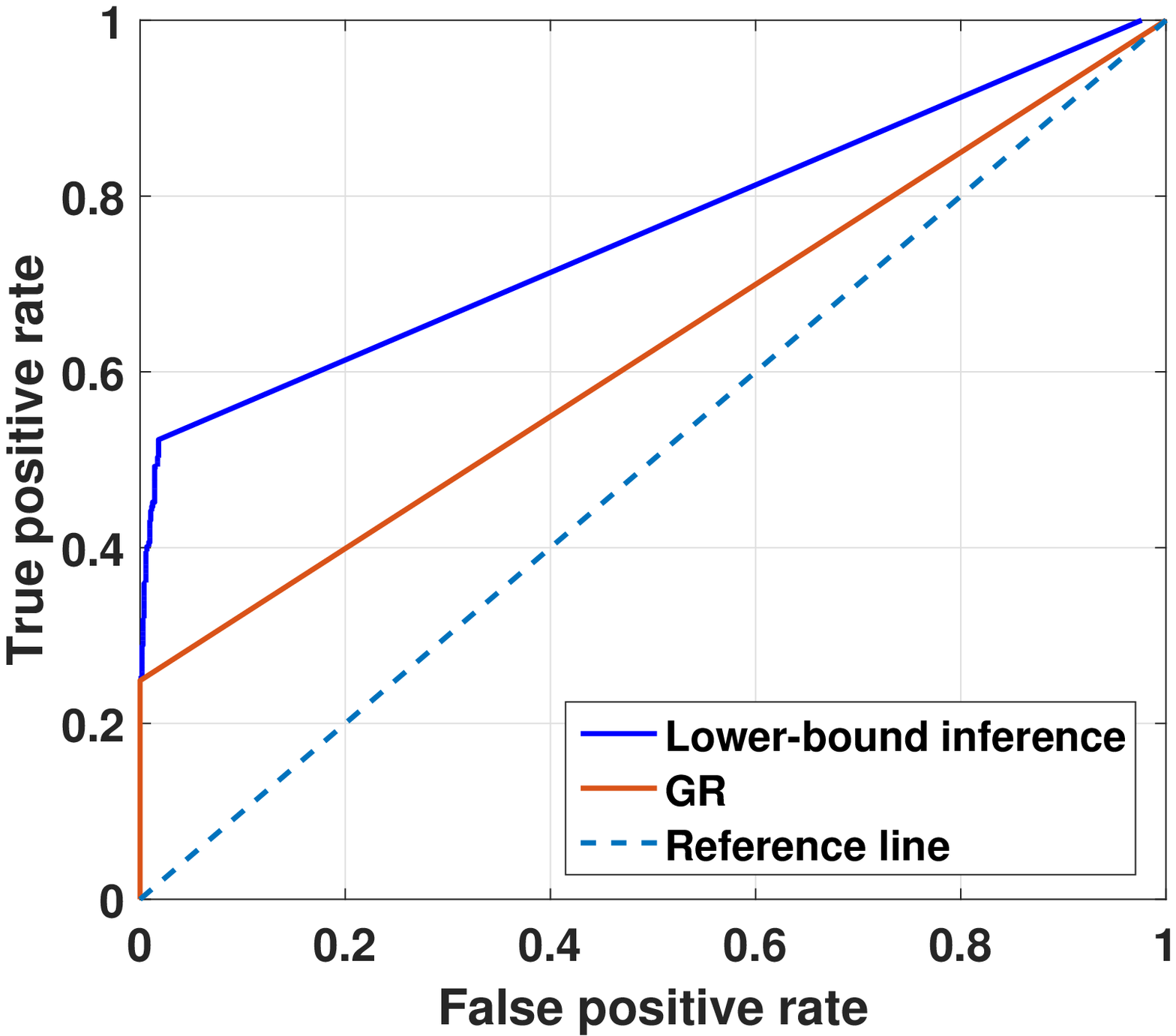}
	}\caption{
	\small
	This figure show an example of the reconstruction results: (a) ROC curves for upper-bound inference and $G_{R}$.
	(b) ROC curves for lower-bound inference and $G_{R}$.}
\end{figure}

\begin{figure*}[thb]

\subfloat[\label{fig:area}]{\includegraphics[width=0.243\textwidth]{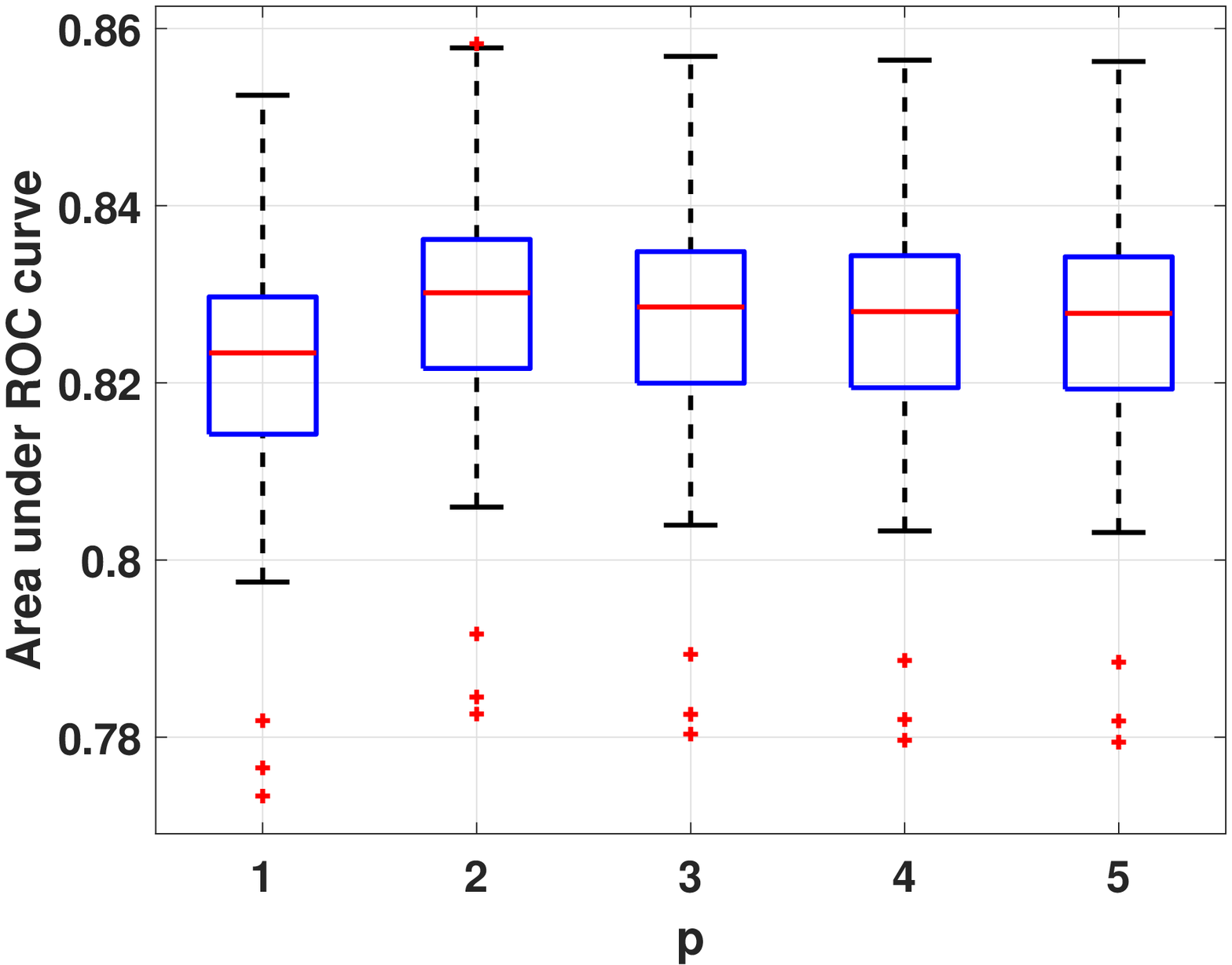} 

}
\subfloat[\label{fig:adv}]{\includegraphics[width=0.243\textwidth]{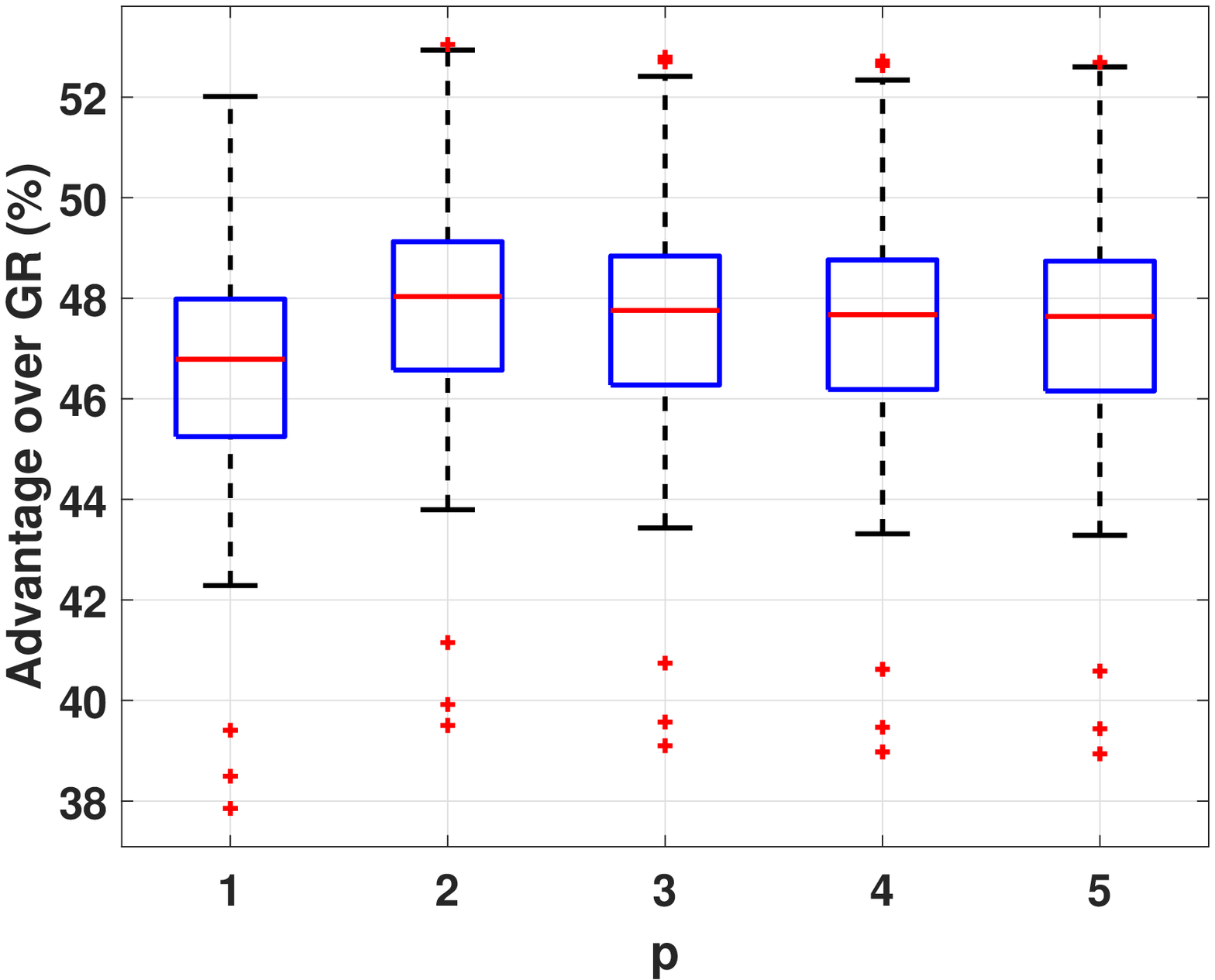}
	
}\subfloat[\label{fig:warea}]{

\includegraphics[width=0.243\textwidth]{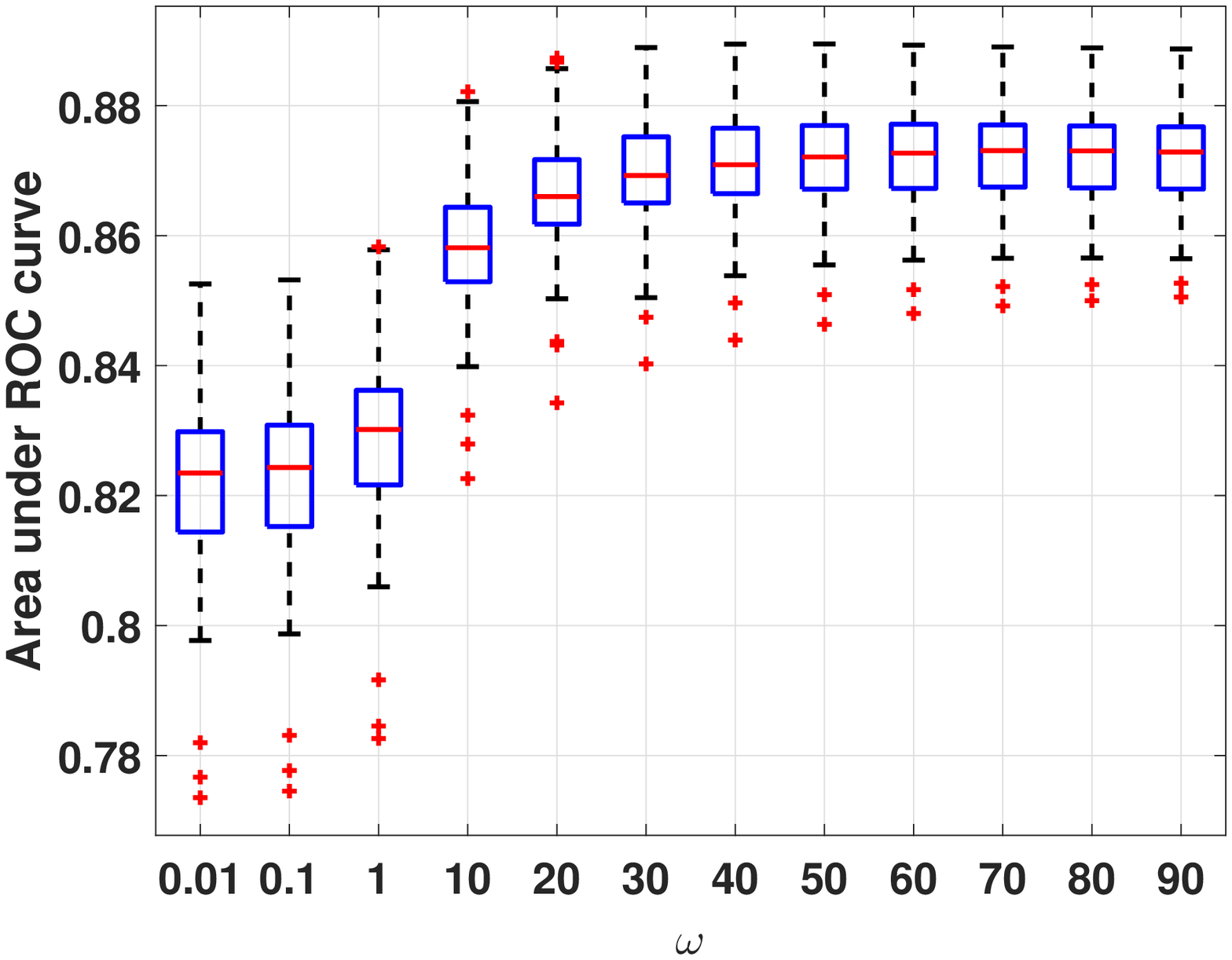}

}\subfloat[\label{fig:wadv}]{

\includegraphics[width=0.243\textwidth]{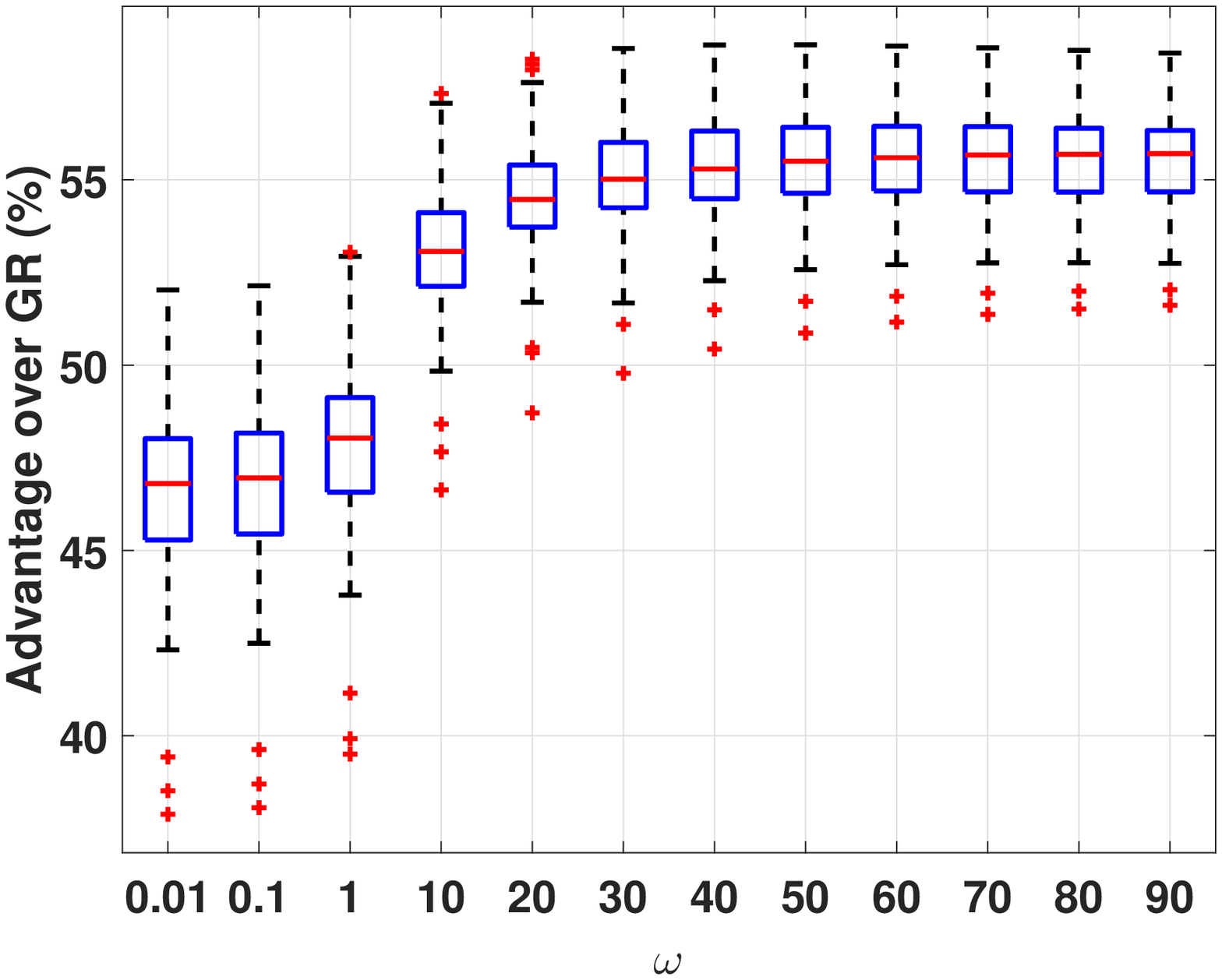}

}

\vspace{-4mm}
\subfloat[\label{fig:edge2}]{
	\includegraphics[width=0.248\textwidth]{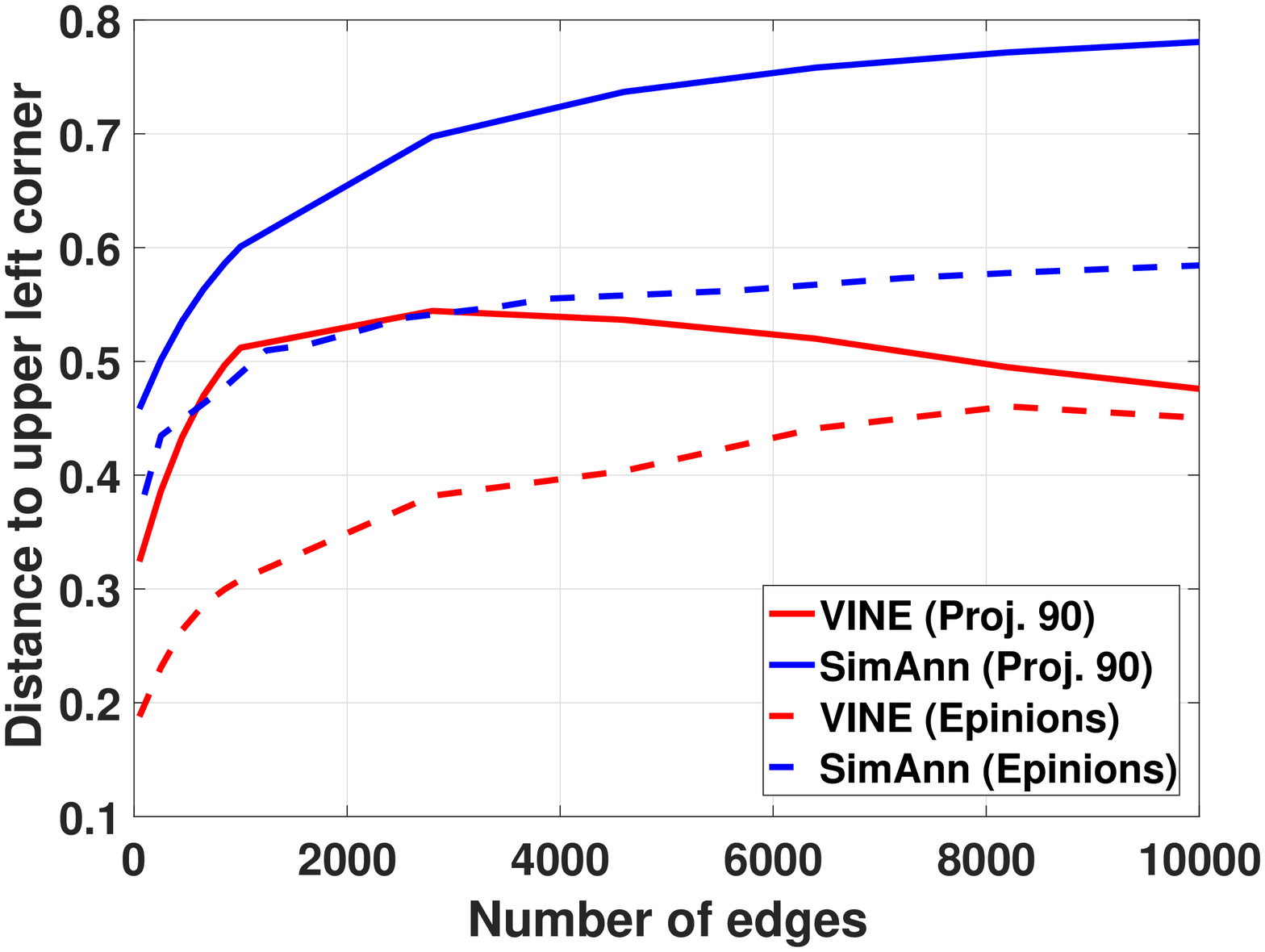}}
\subfloat[\label{fig:time2}]{
	\includegraphics[width=0.248\textwidth]{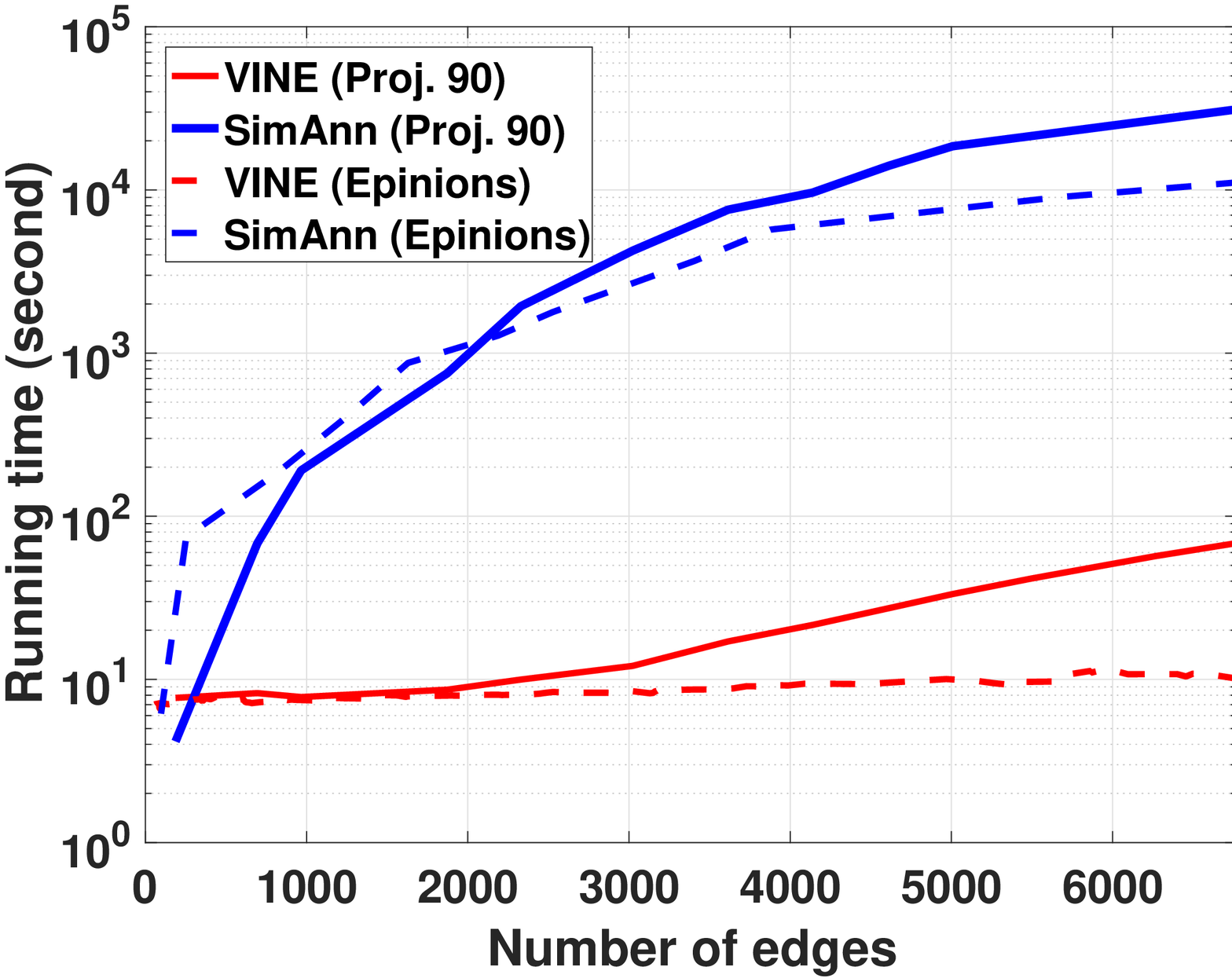}
}
\subfloat[\label{fig:node2}]{
	\includegraphics[width=0.248\textwidth]{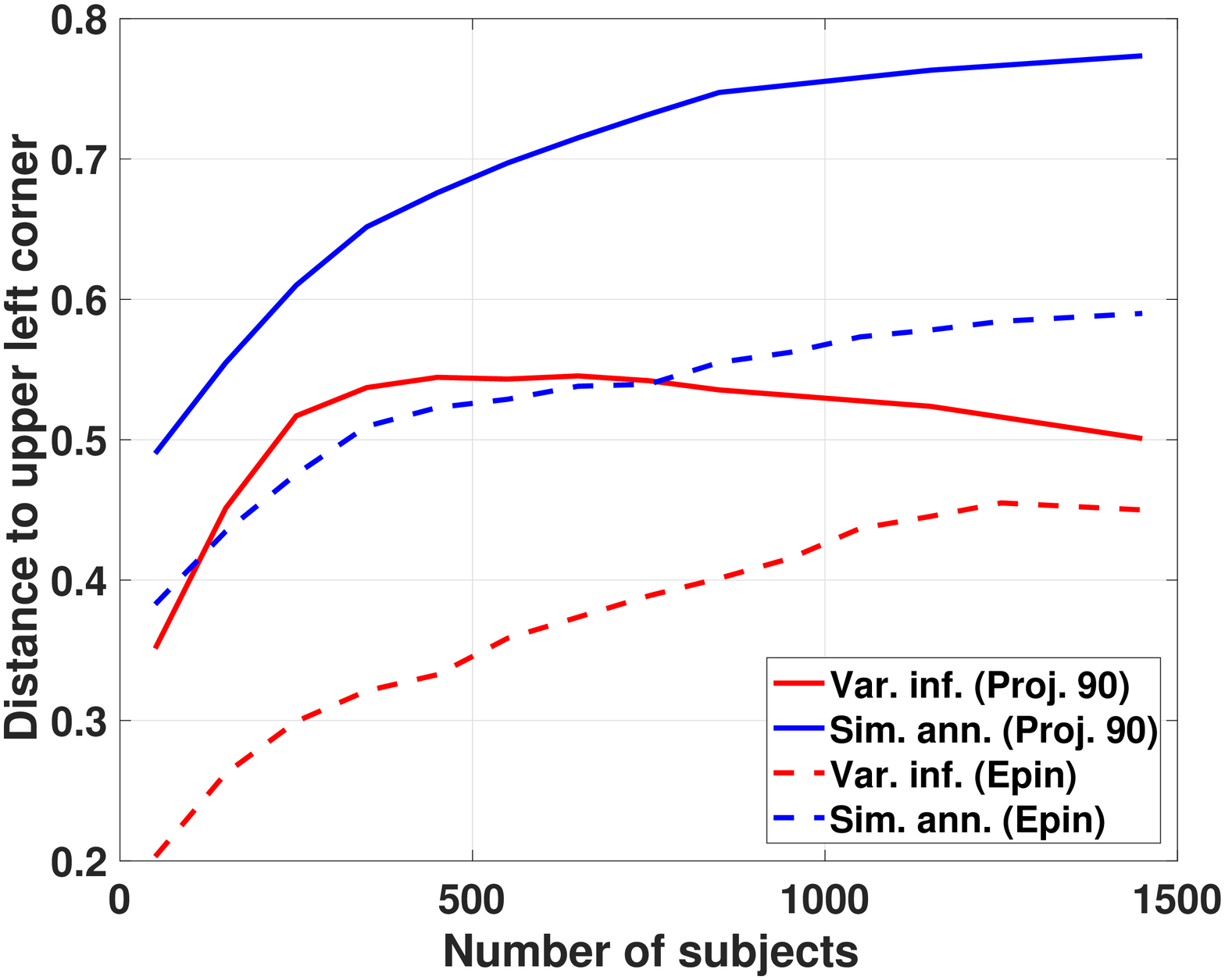}}
\subfloat[\label{fig:seed2}]{
	\includegraphics[width=0.248\textwidth]{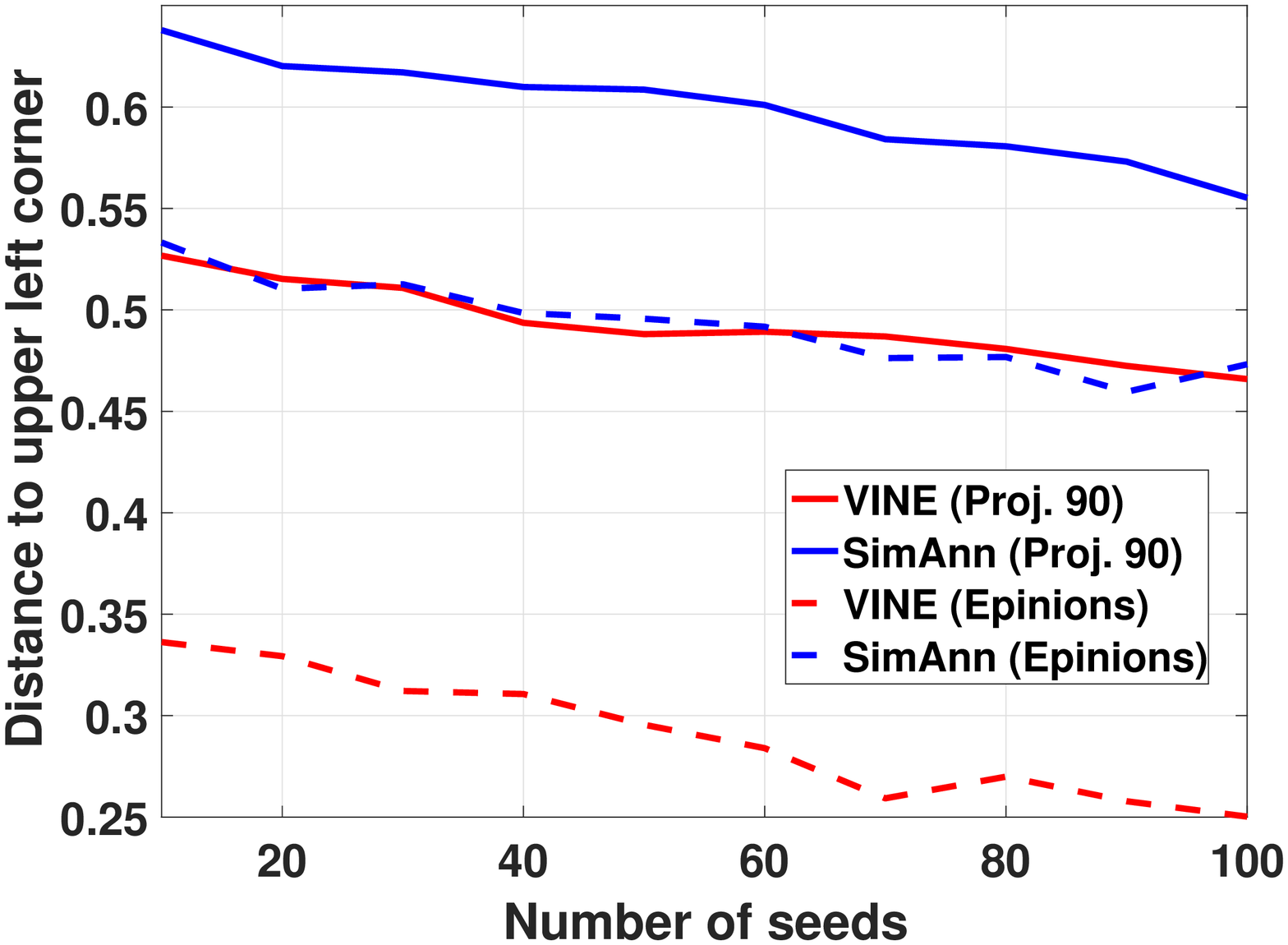}}


\caption{ 
	\small
	(a) The area under the ROC curve of the upper-bound inference, when
	$\psi$ is $L^{p}$ norm and $p$ varies from $1$ to $5$.
	(b)  Improvement of upper-bound inference over $G_{R}$ 
	in terms of the area under the ROC curve, as a function of $p$.
	(c) Area under the ROC curve for upper-bound inference, when we let $\psi$ be
	$\omega\left\Vert \cdot\right\Vert _{2}$ and $\omega$ varies from
	$0.01$ to $100$. (d)  Improvement of upper-bound inference
	over $G_{R}$ in terms of the area under the ROC curve,
	when we let $\psi$ be $\omega\left\Vert \cdot\right\Vert _{2}$ and
	$\omega$ varies from $0.01$ to $100$.
	 (e) Distance to the upper left corner (the smaller, the better) versus the number of edges in $ G_S $. (f) Running time (in seconds) versus the number of edges in $ G_S $. (g) Distance to the upper left corner versus the number of subjects in the sample (the number of nodes in $ G_S $). (h) Distance to the upper left corner (the smaller, the better) versus the number of seeds.
	 }
\end{figure*}

In this section, we evaluate the proposed variational inference algorithm
via experimental results. By varying $\zeta$ from $\min_{i}s_{i}^{\alpha}$
to $\max_{i}s_{i}^{\alpha}$, we obtain a series of inferred adjacency
matrices $\hat{\a}(\zeta)$. Suppose that the true adjacency matrix
of $G_{S}$ is $\a$. The reconstruction performance of an inferred
adjacency matrix $\hat{\a}$ is measured by the true positive rate
(TPR) and the false positive rate (FPR), which are defined as 
$
\mathrm{TPR}(\hat{\a},\a)={\binom{n}{2}}^{-1}\sum_{i<j}1\{\hat{\a}_{ij}=1\mbox{ and }\a_{ij}=1\}
$ and 
$
\mathrm{FPR}(\hat{\a},\a)={\binom{n}{2}}^{-1}\sum_{i<j}1\{\hat{\a}_{ij}=1\mbox{ and }\a_{ij}=0\},
$
where $n$ is the number of subjects. We plot the TPR and FPR of each
$\hat{\a}(\zeta)$ on the ROC plane and obtain the ROC curve. 
\cref{fig:ex,fig:ex_low} show an example of the reconstruction
result. In this example, we simulated an
RDS
process over a real-world network, the Project 90 graph that
represents the community structure of heterosexuals at high risk for HIV infection \cite{woodhouse1994mapping}, 
with inter-recruitment time distribution $\mathrm{Exp}(1)$
(exponential distribution with rate $1$). In the simulation, we choose $n=|V_S|=50$, and a single seed subject at the initial stage; each subject is given 3 coupons;
 1176 missing edges are to be inferred.
 In practice,
the sample size
  $ n $ is usually fixed in advance (according to researchers' study plan).
In Fig. \ref{fig:ex}, the blue ROC curve corresponds to the upper-bound inference. We choose
$\psi$ to be the $L^{2}$ norm. The red curve is a baseline reconstruction given
by estimating $G_S$ by $G_{R}$.  Since $G_{R}$ must be a subgraph of $G_{S}$, the FPR of $G_{R}$ is zero.
The red curve is obtained by connecting the point of the TPR of $G_{R}$
on the vertical axis and the point $(1,1)$. The performance  is quantified by the area of the region under the ROC curve (note that larger is better).
In this example, the area of the region under the blue curve
is $0.92$ and that of the red curve is $0.64$. The region under
the blue curve is $47\%$ greater than that of the red curve. With the best thresholding, the algorithm can achieve a TPR of 90\% while the FPR is only 10\%.
In Fig. \ref{fig:ex_low}, the blue curve is the ROC curve of the
lower-bound inference. We choose $\psi$ to be the $L^{2}$ norm.
The red curve is a baseline given by $G_{R}$. The area under the
blue curve is $0.74$, which is $18\%$ greater than that of the red
curve. Since the lower-bound approximation is obtained from the greedy
algorithm while the upper-bound approximation is the solution to an
optimization problem, we focus on the upper-bound inference.

\subsection{Experiments on Facebook network}
Recall that in \cref{eq:pi}, $\psi$ can be any non-decreasing
convex function. We may let $\psi(\cdot)=\omega\left\Vert \cdot\right\Vert _{p}$ be $\omega$ times the $L^{p}$
norm ($\omega\geq0$, $1\leq p\leq\infty$).
 We now study the influence of different
choices of $\psi$.

\textbf{Influence of $p$.} First we fix $\omega=1$
 and vary $p$ from $1$ to $5$. We simulated $100$
RDS processes over the Facebook network~\cite{mcauley2012learning}. For each $p$, we measure
the area under the ROC curve of the upper-bound inference for each
RDS data and illustrate their distribution with a Tukey boxplot, shown in 
Fig. \ref{fig:area}. We also record the advantage of the area under the ROC curve of the upper-bound
inference over that of $G_{R}$; boxplots are given
in Fig. \ref{fig:adv}. 
 We
observe that the variational inference algorithm achieves remarkably
high accuracy when $p=2,3,4,5$. 

\textbf{Influence of $\omega$.} 
 We let $\psi$ be $\omega\left\Vert \cdot\right\Vert _{2}$
and vary $\omega$ from $0.01$ to $100$.
  $100$ RDS
processes  are simulated over the Facebook network. For each $\omega$, we measure
the area under the ROC curve of the upper-bound inference for each
RDS data and illustrate their distribution with a Tukey boxplot. The
result is presented in Fig. \ref{fig:warea}. In addition, we also
record the advantage of the area under the ROC curve of the upper-bound
inference over that of $G_{R}$. Accordingly, their boxplots are presented
in Fig. \ref{fig:wadv}. From \cref{fig:warea,fig:wadv},
we can observe that the variational inference algorithm achieves higher
accuracy as $\omega$ increases from $0.01$ to $10$ and that the
increase of $\omega$ from $10$ to $100$ leads to lower accuracy
of the variational inference.

\subsection{Experiments on large Project 90 and Epinion networks}\label{sub:p90}
We compare \Alg with the simulated-annealing-based method  proposed in~\cite{chen2015aaai} (referred to as \AlgSA). Since \AlgSA only gives a single point on the ROC plane rather than a curve, thus the reconstruction performance in this set of experiments is quantified by the
  distance from the output point to the upper left corner,
 $ [(1-\mathrm{TPR})^2+\mathrm{FPR}^2]^{1/2}. $ The algorithm with smaller distance
  is considered to attain better performance.
 We  apply both methods to the large Project 90 network~\cite{woodhouse1994mapping} and the Epinions social network~\cite{richardson2003trust}. 
The Epinions network characterizes the trust relationships between users of a general consumer review site Epinions.com.
 
 Fig.~\ref{fig:edge2} shows the distance to the upper left corner versus the number of edges in $ G_S $, where  the $ \zeta $ value is chosen to minimize the distance of the \Alg ROC curve datapoint to the upper-left corner $(0,1)$; we vary from tens of edges to $ 10000 $ edges.
  We generate many RDS processes with different sample sizes and sort them according to the number of edges in $ G_S $ and see the reconstruction performance on these datasets. In this way, we plot how the reconstruction performance varies with the number of edges in $ G_S $. Fig.~\ref{fig:node2} presents the distance to the upper left corner versus the number of subjects in the sample (the number of nodes in $ G_S $).
 \Alg outperforms \AlgSA significantly on both datasets.
  Fig.~\ref{fig:time2} presents the running time (in seconds) versus of the number of edges in $ G_S $. 
   \AlgSA was implemented in C++ while \Alg was written in the Julia language and can be implemented as a parallelized version.
    \Alg runs three orders of magnitude faster than \AlgSA when there are more edges in $ G_S $ and  \Alg is more scalable for large graphs.
 
 Fig.~\ref{fig:seed2} shows the distance to the upper left corner versus the number of seeds. 
  We vary the number of seeds from 10 to 100, while fixing the sample size.
     Both algorithms achieve better reconstruction performance with more seeds and that under the same number of seeds, \Alg attains a remarkably better performance than \AlgSA.
 

\begin{figure}[htb]
	\centering
	\includegraphics[width=0.4\columnwidth]{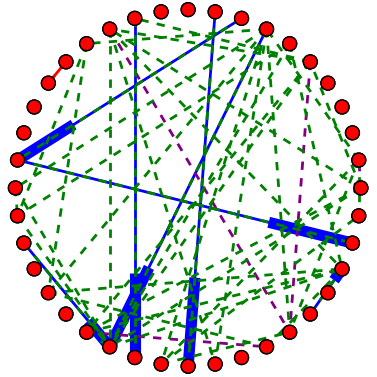}
	\caption{
		\small
		This figure illustrates the reconstruction result for a subnetwork of size 40 of the Epinions network. Blue arrows denote the direct recruitment links revealed by $G_R$ (only 6); green edges are the correctly inferred links; purple edges are the incorrectly inferred links;  only one red edge is missed by our proposed method. {In this example, $ 45 $ out of $ 46 $ edges (approximately $ 97.8 \% $) are successfully recovered.}\label{fig:illu}}
\end{figure}
  
  Fig.~\ref{fig:illu} visualizes the reconstruction result for a subnetwork of size 40 of the Epinions network. Only 6 recruitment links are revealed in $ G_R $.
     $ 45 $ out of $ 46 $ edges (approximately $ 97.8 \% $) are successfully recovered.

\clearpage{}
\bibliographystyle{abbrv}
{
	
	\bibliography{reference-list}
}

\clearpage{}

\appendix
\section*{Appendix}
\renewcommand{\thesubsection}{\Alph{subsection}}


\subsection{Likelihood of Recruitment Time Series}\label{sub:likeli}

	
	We consider the recruitment of subject $i$. Recall that $R_{u}(i)$ denotes
	the set of recruiters of subject $u$ just before time $\tt_{i}$ and that
	$I_{u}(i)$ denotes the set of potential recruitees of recruiter $u$ just before
	time $\tt_{i}$. 
	
	We compute the likelihood of the $ i $-th recruitment event (the recruitment of subject $ i $) in the two cases: $ i $ enters the study via the recruitment of a subject already in the study (in this case, subject $ i $ is not a seed node, which is denoted by $ i\notin M $) and via the direct recruitment of the researchers (in this case, subject $ i $ is a seed node, which is denoted by $ i\in M $).
	
	Suppose that $i\notin M$. The inter-recruitment time
	between $i$ and its potential recruiter $u$ is denoted $W_{ui}=\tt_{i}-\tt_{u}$
	and is greater than $\tt_{i-1}-\tt_{u}$ conditional on previous recruitment
	of $i$. Let $U$ be the random variable of next recruiter and $X$
	be the random variable of next recruitee, namely the subject that will be labeled as subject $ i $. We would like to note here that subject $ i $ is in fact random. Let $J$ denote the event
	$\forall j\in R(i),k\in I(i),W_{jk}>\tt_{i-1}-\tt_{j}$. 
	
	We first compute the probability that a certain subject $ x \in I(i) $ is the next ($ i $-th) recruitee, $ u \in R_x(i) $ is its recruiter, and the inter-recruitment time between $ u $ and $ x $ is greater than or equal to $ t - \tt_u $, conditional on event $ J $. Intuitively, $ t $ is the recruitment time of subject $ x $ and in fact we are computing the tail probability of $ W_{ux} $. We condition on the event $ J $ because having observed the $ (i-1) $-th recruitment event, we know that for all possible recruiter-recruitee pairs in the next ($ i $-th) recruitment event, say $ j\in R(i) $ and $ x\in I(i) $, their inter-recruitment time $ W_{jk} $ should be greater than or equal to $ \tt_{i-1}-\tt_{j} $ (otherwise, the event that subject $ j $ recruits subject $ x $ will happen before $ \tt_i $ and they will not appear in $ R(i) $ and $ I(i) $, respectively). 
	We have
	\begin{align}\label{eq:first_eq}
		& \p\left[U=u,X=x,W_{ux}\geq t-\tt_{u}\mid J\right]\nonumber\\
		= & \p\left[W_{ux}\geq t-\tt_{u},\tt_{j}+W_{jk}>\tt_{u}+W_{ux},\right.\nonumber\\
		& \left.\forall j\in R(i),k\in I(i),\{u,x\}\neq\{j,k\}\mid J\right].
	\end{align}
	Since the $ i $-th recruitment event is that $ u $ recruits $ x $, the inter-recruitment time along this link must be minimum among those along all other links. Therefore, in \cref{eq:first_eq} we consider $ W_{jk} $ for $\forall j\in R(i),k\in I(i),\{u,x\}\neq\{j,k\}$. We require that $$\tt_{j}+W_{jk}>\tt_{u} +W_{ux},$$
	which means exactly that the recruitment time of $ x $ ($ \tt_{u} +W_{ux} $) is minimum (smaller than the recruitment time of $ k $ for all $ k $). 
	
	Then we marginalize the above probability in \cref{eq:first_eq} over all possible combinations of $ x $ and $ u $.
	Recall that any subject in $ x\in I(i) $ could possibly be the subject $ i $ and any subject in $ R_x(i) $ could be her recruiter; therefore we need to sum over all possible recruitee-recruiter combinations, i.e., sum over $x\in I(i)$ and $ u\in R_x(i) $:
	\begin{eqnarray}
		&  & \p\left[W_{Ui}\geq t-\tt_{U}\mid J\right]\nonumber\\
		& = & \sum_{x\in I(i)}\sum_{u\in R_{x}(i)}\p\left[W_{ux}\geq t-\tt_{u},\tt_{j}+W_{jk}>\tt_{u}\right.\nonumber\\
		&  & \left.+W_{ux},\forall j\in R(i),k\in I(i),\{u,x\}\neq\{j,k\}\mid J\right]\nonumber\\
		& = & \sum_{x\in I(i)}\sum_{u\in R_{x}(i)}\int_{t-\tt_{u}}^{\infty}\rho_{\t(u;i)}(s)\d s\p\left[W_{jk}>s+\tt_{u}\right.\nonumber\\
		&  & \left.-\tt_{j},\forall j\in R(i),k\in I(i),\{u,x\}\neq\{j,k\}\mid J\right]\nonumber\\
		& = & \sum_{x\in I(i)}\sum_{u\in R_{x}(i)}\int_{t-\tt_{u}}^{\infty}\rho_{\t(u;i)}(s)\cdot\nonumber\\
		&  & \frac{\prod_{j\in R(i)}\prod_{k\in I_{j}(i)}\left(1-D_{\t(j;i)}(s+\tt_{u}-\tt_{j})\right)}{1-D_{\t(u;i)}(s)}\d s\nonumber\\
		& = & \sum_{x\in I(i)}\sum_{u\in R_{x}(i)}\int_{t-\tt_{u}}^{\infty}H_{\t(u;i)}(s)\cdot\nonumber\\
		&  & \prod_{j\in R(i)}S_{\t(j;i)}^{|I_{j}(i)|}(s+\tt_{u}-\tt_{j})\d s.\nonumber
	\end{eqnarray}
	Using the notation that we introduced in Section~\ref{sec:Network-Reconstruction-for} to re-write and simplify the above expression, we obtain the likelihood of the $ i $-th recruitment event for $ i\notin M $:
	\begin{align*}
		& \sum_{x\in I(i)}\sum_{u\in R_{x}(i)}H_{\t(u;i)}(\tt_{i}-\tt_{u})\prod_{j\in R(i)}S_{\t(j;i)}^{|I_{j}(i)|}(\tt_{i}-\tt_{j})\\
		= & \prod_{j\in R(i)}S_{\t(j;i)}^{|I_{j}(i)|}(\tt_{i}-\tt_{j})\sum_{u\in R(i)}|I_{u}(i)|H_{\t(u;i)}(\tt_{i}-\tt_{u})
	\end{align*}
	Now suppose that $i\in M$, which means that subject $ i $ is recruited into the study directly by the researchers. Therefore the inter-recruitment time of any possible recruiter-recruitee pairs of the $ i $-th recruitment event, say $ j $ and $ k $, should be greater than or equal to $ t-\tt_j $, where $ t $ is the recruitment time of subject $ i $. In the terminology of survival analysis, all these potential recruitment links are censored. So we compute the probability that all these links are censored:
	\begin{align*}
		& \p\left[W_{jk}\geq t-\tt_{j},\forall j\in R(i),k\in I_{j}(i)\mid J\right]\\
		= & \prod_{j\in R(i)}\prod_{k\in I_{j}(i)}(1-D_{\t(j;i)}(t-\tt_{j}))\\
		= & \prod_{j\in R(i)}S_{\t(j;i)}^{|I_{j}(i)|}(t-\tt_{j}).
	\end{align*}
	Plugging in the observed recruitment time of subject $ i $ (denoted by $ \tt_i $), we obtain the likelihood of the $ i $-th recruitment event for $ i\in M $: 
	\[
	\prod_{j\in R(i)}S_{\t(j;i)}^{|I_{j}(i)|}(\tt_{i}-\tt_{j}).
	\]
	So far we have obtained the likelihood of the $ i $-th recruitment event for both cases ($ i\notin M $ and $ i\in M $); Multiplying the likelihoods with $ i $ running from $ 1 $ to $ n $, we have 
	the entire likelihood:
	\begin{multline}\label{eq:likeli}
		\prod_{i=1}^{n}\left(\sum_{u\in R(i)}|I_{u}(i)|H_{\t(u;i)}(\tt_{i}-\tt_{u})\right)^{1\{i\notin M\}}\cdot\\
		\prod_{j\in R(i)}S_{\t(j;i)}^{|I_{j}|}(\tt_{i}-\tt_{j}),
	\end{multline}
	where $ 1\{i\notin M\} $ the indicator random variable for the event that $ i\notin M $.
	%

\subsection{Log-likelihood of Recruitment Time Series}\label{sub:logl}
	According to \cref{eq:likeli} in 
	 Appendix~\ref{sub:likeli}
	, the log-likelihood is
	\begin{multline*}
	\sum_{i=1}^{n}\left[1\{i\notin M\}\log\left(\sum_{u\in R(i)}|I_{u}(i)|H_{\t(u;i)}(\tt_{i}-\tt_{u})\right)\right.\\
	\left.+\sum_{j\in R(i)}|I_{j}(i)|\log S_{\t(j;i)}(\tt_{i}-\tt_{j})\right].
	\end{multline*}
	The number of recruitees of recruiter $u$ just before time $\tt_{i}$
	is given by
	\begin{equation}
	|I_{u}(i)|=\c_{ui}\left(\sum_{k=i}^{n}\a_{uk}+\u_{u}\right).\label{eq:I_j(i)}
	\end{equation}
	The cardinality of $ I_{u}(i) $ is zero if and only if recruiter $ u $ has at least one coupon just before $ \tt_i $, i.e.,  $ \c_{ui} = 1$. Therefore there is a factor $ \c_{ui} $ in \cref{eq:I_j(i)}. When recruiter $ u $ has at least one coupon, the number of recruitees of recruiter $u$ just before time $\tt_{i}$ is \[ \sum_{k=i}^{n}\a_{uk}+\u_{u}, \]
	where $\sum_{k=i}^{n}\a_{uk}  $ is the number of recruitees in the final sample and $ \u_{u} $ is the number of recruitees outside the final sample. In the expression $ \sum_{k=i}^{n}\a_{uk} $, we sum over $ k $ from $ i $ to $ n $ since subjects $ i,i+1,\ldots,n $ are those in the final sample and recruited at and after time $ \tt_i $ and they contribute one to the sum if they are adjacent to subject $ u $ (namely $ \a_{uk} =1 $).
	
	Recall that $ \b $ is the Hadamard product of $ \c $ and $ \h $, which yields that $ \b_{ui} = \c_{ui} \h_{ui} $, and that $$\h_{ui}=H_{\t(u;i)}(\tt_{i}-\tt_{u}).$$ Therefore the term $$\sum_{u\in R(i)}|I_{u}(i)|H_{\t(u;i)}(\tt_{i}-\tt_{u})$$ in
	the log-likelihood can be written as
	
	\begin{align*}
	& \sum_{u\in R(i)}|I_{u}(i)|H_{\t(u;i)}(\tt_{i}-\tt_{u})\\
	= & \sum_{u}\c_{ui}\left(\sum_{k=i}^{n}\a_{uk}+\u_{u}\right)\h_{ui}\\
	= & (\b'\u+\lt(\a\b)'\cdot\mathbf{1})_{i}.
	\end{align*}
	
	Recall the definition of the matrix $ \s $: $$\s_{ji}=\log S_{\t(j;i)}(\tt_{i}-\tt_{j}),$$ and that $ \D $ is the Hadamard product of $ \c $ and $ \s $, which yields that $ \D_{ji} = \c_{ji}\s_{ji} $. Similarly, the term $$\sum_{j\in R(i)}|I_{j}(i)|\log S_{\t(j;i)}(\tt_{i}-\tt_{j})$$
	in the log-likelihood is given by
	
	\begin{align*}
	& \sum_{j\in R(i)}|I_{j}(i)|\log S_{\t(j;i)}(\tt_{i}-\tt_{j})\\
	= & \sum_{j\in R(i)}\c_{ji}\left(\sum_{k=i}^{n}\a_{jk}+\u_{j}\right)\s_{ji}\\
	= & (\D'\u+\lt(\a\D)'\cdot\mathbf{1})_{i}.
	\end{align*}

	Thus the log-likelihood is
	\begin{multline*}
	\sum_{i=1}^{n}\left[1\{i\notin M\}\log\left(\b'\u+\lt(\a\b)'\cdot\mathbf{1}\right)_{i}\right.\\
	\left.+\left(\D'\u+\lt(\a\D)'\cdot\mathbf{1}\right)_{i}\right]=\mathbf{m}'\beta+\mathbf{1}'\delta.
	\end{multline*}
	

\subsection{Proof of Theorem \ref{thm:log-submodular}}
\label{sub:proofthm}
In this section, we will show that $ \log \tilde{L}(\g) $ is submodular in $ \g $.
 We have
\begin{eqnarray*}
&& \log \tilde{L}(\g)\\
 & = & \log L(\tt|\a,\theta)+\log\pi(\a)+\log\phi(\theta)\\
 & = & \mathbf{m}'\beta+\mathbf{1}'\delta-\psi(\max\{\u+\a\cdot\mathbf{1}-\dd,\mathbf{0}\}\})+\log\phi(\theta).
\end{eqnarray*}
Later we will show that it is submodular part by part.
 
First, we need to prove that $ -\psi(\max\{\u+\a\cdot\mathbf{1}-\dd,\mathbf{0}\}\}) $ is submodular. 
We temporarily view them as real-valued vectors and matrices rather
than binary vectors and matrices. In light of \cref{eq:u-mu},
we know that $\u$ is a linear function of $\bm{\mu}$, which yields that $ \u+\a\cdot\mathbf{1}-\dd $ is a linear function of $ \bm{\mu} $. We know that if $ g(\x) $ is a linear function, then $ f(\x) = \max \{ g(\x), 0\}$ is a convex function (see Section~3.2.3 in~\cite{boyd2004convex}). Therefore, every entry
of $\max\{\u+\a\cdot\mathbf{1}-\dd,\mathbf{0}\}$ is convex in $\u$.
 Since $\psi$ is a convex function
and non-decreasing in each argument whenever this argument is non-negative,
thus $\psi(\max\{\u+\a\cdot\mathbf{1}-\dd,\mathbf{0}\}\})$ is convex
in $\bm{\u}$ (see Section~3.2.4 in~\cite{boyd2004convex}); equivalently, $-\psi(\max\{\u+\a\cdot\mathbf{1}-\dd,\mathbf{0}\}\})$
is concave in $\bm{\u}$. Thus this term $ -\psi(\max\{\u+\a\cdot\mathbf{1}-\dd,\mathbf{0}\}\}) $ is submodular in $ \bm{\mu} $ if we view $ \bm{\mu} $ as a Boolean vector.

Recall the definitions of $ \beta $ and $ \delta $:
\begin{eqnarray*}
	\beta&=&\log(\b'\u+\lt(\a\b)'\cdot\mathbf{1}),\\
	\delta&=&\D'\u+\lt(\a\D)'\cdot\mathbf{1}.
\end{eqnarray*}
 The function $\beta(\u,\a)$ is concave
in $\u$ and $\a$ since the inner part $$ \b'\u+\lt(\a\b)'\cdot\mathbf{1} $$ is linear in $ \u $ and $ \a $, the logarithm function is concave, and $ \beta $ is the composition of the linear inner part and the concave logarithm function. The function $\delta(\u,\a)$ is linear in
$\u$ and $\a$. 
Recall that $ \u $ and $ \a $ are linear in $ \bm{\mu} $ and $ \bm{\alpha} $, respectively.
Thus $\beta$ is concave in $\bm{\mu}$ and $\bm{\alpha}$
and $\delta$ is linear in $\bm{\mu}$ and $\bm{\alpha}$. Therefore
$\beta$ is concave in $\g$ and $\delta$ is linear in
$ \g $, where $$\g=(\aa,\bm{\mu}).$$ Thus
$\mathbf{m}'\beta+\mathbf{1}'\delta$ is submodular in $\g$ if $ \g $ is viewed as a binary vector. 

Hence in light of the fact that the sum of submodular functions is submodular, the
whole expression is submodular in $ \g $. In other words,  $\log L(\g)$
is submodular in $\g$.

\subsection{Proof of Proposition~\ref{prop:lower}}\label{sub:lower}

We prove it by induction.

Suppose that $\g = \{v_{j_1},v_{j_2},\ldots,v_{j_q}\}$, where $j_1< j_2< \cdots < j_q$ and $q\leq N$.

If $q=1$, then 
\begin{multline*}
F(\g) - F(\varnothing) = F(\{v_{j_1}\}) - F(\varnothing)\\ \geq F(V_{j_1-1}\cup v_{j_1}) - F(V_{j_1-1}) = s^g(v_{j_1}) = s^g(\g),
\end{multline*}
since $\varnothing$ must be a subset of $V_{j_1-1}$. Therefore, $$F(\g)\geq s^g(\g)+F(\varnothing) = s^g(\g),$$ since $F$ is normalized.

Suppose that the proposition holds for all $q<r$. When $q=r$, we have
\begin{multline*}
F(\{ v_{j_1},v_{j_2},\ldots,v_{j_r} \}) - F(\{ v_{j_1},v_{j_2},\ldots,v_{j_{r-1}} \})\\
 \geq F(V_{j_{r}-1} \cup \{v_{j_r}\}) - F(V_{j_{r}-1}) = s^g(\{v_{j_r}\}), 
\end{multline*}
since $\{ v_{j_1},v_{j_2},\ldots,v_{j_{r-1}} \}$ is a subset of  
$V_{j_{r}-1}$. Therefore, we obtain
\begin{equation}
\begin{split}
F(\g) & = F(\{ v_{j_1},v_{j_2},\ldots,v_{j_r} \})\\
& \geq s^g(\{v_{j_r}\}) + F(\{ v_{j_1},v_{j_2},\ldots,v_{j_{r-1}} \})\\
& \geq s^g(\{v_{j_r}\}) + s^g(\{ v_{j_1},v_{j_2},\ldots,v_{j_{r-1}} \})\\
& = s^g(\{ v_{j_1},v_{j_2},\ldots,v_{j_r} \})\nonumber
\end{split}
\end{equation}
by the induction assumption. This completes the proof.

\subsection{Proof of Proposition~\ref{prop:sup}}\label{sub:sup}
In order to show that a modular function $s$ is a supergradient of the submodular function $F$ at $\x$, we have to show that
\[\forall{\bf y}\in\{0,1\}^{N},F({\bf y})\leq F({\bf x})+s({\bf y})-s({\bf x}).\]
Equivalent, if viewed as a set function, we have to show that
\[\forall Y \subseteq [N],F(Y)\leq F(X)+s(Y)-s(X),\]
where $X$ is the corresponding subset for $\x$, i.e., \[X=\{i\in [N]:\x_i=1\}.\]
Since $s$ is a modular function, it is equivalent to show \[F(Y) + \sum_{i\in X\setminus Y} s(\{i\})\leq F(X)+ \sum_{i\in Y\setminus X} s(\{i\})  .\]
\paragraph{Grow supergradient}
We have to show that
\[F(Y) + \sum_{i\in X\setminus Y} \hat{s}(\{i\})\leq F(X)+ \sum_{i\in Y\setminus X} \hat{s}(\{i\});\]
equivalently,
\[F(Y) + \sum_{i\in X\setminus Y} \Delta_i F(V-\{i\})\leq F(X)+ \sum_{i\in Y\setminus X} \Delta_i F(X).\]
We will show that the left-hand side is less than or equal to $F(X\cup Y)$ while the right-hand side is greater than or equal to $F(X\cup Y)$.

Suppose that $Y\setminus X = \{a_1,a_2,a_3,\ldots,a_r\}$, $A_i = X\cup \{a_1,a_2,a_3,\ldots,a_i\}$ and $A_0=X$. We have \begin{equation}
\begin{split} 
F(X\cup Y) & =  F(X)+ \sum_{i=1}^r (F(A_i) - F(A_{i-1})  ) \\
 &= F(X) + \sum_{i=1}^r \Delta_{a_i} F(A_{i-1})\\
& \leq F(X) + \sum_{i=1}^r \Delta_{a_i} F(X) \\
&=  F(X) + \sum_{i\in Y\setminus X} \Delta_i F(X).
\label{eq:first}
\end{split}
\end{equation}
Suppose that $X\setminus Y = \{ b_1,b_2,\ldots,b_q \}$, $B_i = Y\cup \{b_1,b_2,\ldots,b_i\}$ and $B_0=Y$. We have
\begin{equation}
\begin{split} F(X\cup Y) & = F(Y)+ \sum_{i=1}^q (F(B_i) - F(B_{i-1})  ) \\
& = F(Y) + \sum_{i=1}^q \Delta_{b_i} F(B_{i-1})\\
& \geq F(Y) + \sum_{i=1}^q \Delta_{b_i} F(V-\{b_i\}) \\
& = F(Y) + \sum_{i\in X\setminus Y} \Delta_i F(V-\{i\}).
\label{eq:second}
\end{split}
\end{equation}
\paragraph{Shrink supergradient}
We have to show that
\[F(Y) + \sum_{i\in X\setminus Y} \check{s}(\{i\})\leq F(X)+ \sum_{i\in Y\setminus X} \check{s}(\{i\});\]
equivalently,
\[F(Y) + \sum_{i\in X\setminus Y} \Delta_i F(X-\{i\})\leq F(X)+ \sum_{i\in Y\setminus X} F(\{i\}).\]
We will show that the left-hand side is less than or equal to $F(X\cup Y)$ while the right-hand side is greater than or equal to $F(X\cup Y)$.

In light of \cref{eq:first}, we have
\begin{equation}
\begin{split} F(X\cup Y) & = F(X) + \sum_{i=1}^r \Delta_{a_i} F(A_{i-1})\\
& \leq F(X) + \sum_{i=1}^r \Delta_{a_i} F(\varnothing) \\
& = F(X) + \sum_{i\in Y\setminus X} \Delta_i F(\varnothing) \\
& = F(X) + \sum_{i\in Y\setminus X} F(\{i\}).
\label{eq:third}
\end{split}
\end{equation}
In light of \cref{eq:second}, we have
\begin{equation}
\begin{split} F(X\cup Y)& = F(Y) + \sum_{i=1}^q \Delta_{b_i} F(B_{i-1})\\
& \geq F(Y) + \sum_{i=1}^q \Delta_{b_i} F(X-\{b_i\}) \\
& = F(Y) + \sum_{i\in X\setminus Y} \Delta_i F(X-\{i\}).
\nonumber
\end{split}
\end{equation}
\paragraph{Bar supergradient}
We have to show that
\[F(Y) + \sum_{i\in X\setminus Y} \bar{s}(\{i\})\leq F(X)+ \sum_{i\in Y\setminus X} \bar{s}(\{i\});\]
equivalently,
\[F(Y) + \sum_{i\in X\setminus Y} \Delta_i F(V-\{i\})\leq F(X)+ \sum_{i\in Y\setminus X} F(\{i\}).\]
By \cref{eq:second}, we know that the left-hand side is less than or equal to $F(X\cup Y)$. By \cref{eq:third}, we know that the right-hand side is greater than or equal to $F(X\cup Y)$. Therefore, the left-hand side is less than or equal to the right-hand side.

\subsection{Discussion}\label{sec:discuss}

In some ways, RDS resembles a diffusion process and in fact the continuous-time stochastic process that we formulate RDS as in this paper is a diffusion process; but RDS reveals an extra piece of information that makes reconstruction of the induced subgraph possible: the degrees of each vertex visited by the diffusion process.

For other diffusion processes, we can also derive their likelihood functions. If the optimization problem of the likelihood or the posterior is log-submodular and unconstrained, we can use the submodular variational inference method that we used in this paper. If it is a constrained problem, it is natural to relax the constraints with multiplicative log-submodular penalty factor such that the product of the original likelihood/posterior function and the penalty factor remain log-submodular. For example, if the constraint is an equality constraint of the form $ q(\a) = \bm{c}_0 $, where $ q $ is a linear function (e.g., multiplying $ \a $ by some matrix), $ \bm{c}_0 $ is a fixed vector, and $ \a $ is the adjacency matrix to be optimized over, then we can add a multiplicative factor $ e^{-\lVert q(\a) - \bm{c}_0 \rVert } $ to the likelihood or posterior function, in light of the fact that every norm is a convex function, which guarantees that $ e^{-\lVert q(\a) - \bm{c}_0\rVert } $ is log-submodular with respect to $ \a $. If the constraint is given by an inequality, we can mimic the method that we used in Section~\ref{sub:removal} by introducing auxiliary variables $ \u $ and adding a multiplicative penalty term similar to \cref{eq:pi}.

\end{document}